# *ToxTree: descriptor-based machine learning models for both hERG and Nav1.5 cardiotoxicity liability predictions*★

Issar Arab[1,2] * and Khaled Barakat[2]

[1]*Department of Informatics, Technical University of Munich, Garching, Bavaria 85748, Germany*

[2]*Faculty of Pharmacy and Pharmaceutical Sciences, University of Alberta, Edmonton, Alberta 8613, Canada*

A B S T R A C T

*Drug-mediated blockade of the voltage-gated potassium channel(hERG) and the voltage-gated sodium channel (Nav1.5) can lead to severe cardiovascular complications. This rising concern has been reflected in the drug development arena, as the frequent emergence of cardiotoxicity from many approved drugs led to either discontinuing their use or, in some cases, their withdrawal from the market. Predicting potential hERG and Nav1.5 blockers at the outset of the drug discovery process can resolve this problem and can, therefore, decrease the time and expensive cost of developing safe drugs. One fast and cost-effective approach is to use in silico predictive methods to weed out potential hERG and Nav1.5 blockers at the early stages of drug development. Here, we introduce two robust 2D descriptor-based QSAR predictive models for both hERG and Nav1.5 liability predictions. The machine learning models were trained for both regression, predicting the potency value of a drug, and multiclass classification at three different potency cut-offs (i.e. 1µM, 10µM, and 30µM), where ToxTree-hERG Classifier, a pipeline of Random Forest models, was trained on a large curated dataset of 8380 unique molecular compounds. Whereas ToxTree-Nav1.5 Classifier, a pipeline of kernelized SVM models, was trained on a large manually curated set of 1550 unique compounds retrieved from both ChEMBL and PubChem publicly available bioactivity databases. The hERG model yielded a multiclass accuracy of Q4=74.5% and a binary classification performance of Q2 = 93.2%, sensitivity = 98.7%, specificity = 75%, MCC = 80.3%, and a CCR=86.8% on an external test set of N=499 compounds. The proposed inducer outperformed most metrics of the state-of-the-art published model and other existing tools. Additionally, we are introducing the first Nav1.5 liability predictive model achieving a Q4 = 74.9% and a binary classification of Q2 = 86.7% with MCC = 71.2% and F1 = 89.7 % evaluated on an external test set of 173 unique compounds. The curated datasets used in this project are made publicly available to the research community.*

## 1. Introduction

Drug discovery is an immensely expensive and time-consuming process posing numerous formidable challenges. The development of a new drug requires 6 to12 years and costs as much as $2.6 billion [1, 2, 3]. Computer-aided drug discovery (CADD) holds a significant promise to reduce the costs and to speed up the development of lead candidates at the outset of drug discovery [3, 4]. In recent years, toxicity prediction algorithms have attracted growing attention from researchers in academia and industry alike. They represent an increasingly important component of modern CADD [4], with cardiotoxicity prediction algorithms being at the forefront of these methods.

★This research project received funding from the Natural Sciences and Engineering Research Council of Canada (NSERC).
*Corresponding author: issar.arab@tum.de*

The main motivation driving this new direction is the significant losses endured by pharmaceutical companies in withdrawing several drugs from the market or in halting many drug discovery programs in their pipelines due to unforeseen cardiotoxicities during the early stages of drug development. Furthermore, an additional motivation inspiring the field is the exponential increase of bioactivity data on cardiac ion channel blockers deposited in public databases (*e.g.* ChEMBL and PubChem). These huge data pools made cardiac ion channels an excellent avenue to apply modern statistical models to predict their drug off-target interactions and to achieve high performances via the use of Machine Learning (ML) algorithms.

Quantitative Structure-Activity Relationship (QSAR) [5] is a well-established strategy in the field of chemistry and pharmacy for the reliability and reproducibility of the constructed models. As most of the earlier research was focused only on predicting hERG liability and ignored other cardiac ion channels (*e.g.* voltage-gated sodium channel(Nav1.5) and the voltage-gated calcium channel($Ca_v1.2$)), this paper represents a major step towards closing this gap. Here we follow the best practices methodology of QSAR to build an alternative hERG liability prediction model. Our hERG solution competes with Kumar et al. [6] model. While our model is trained on the same large development hERG dataset and tested on the same evaluation set used by Kumar et al. [6], but further cleaned from existing duplicates, it outperformed most of their metrics with higher performances and with more in-depth details of the sub-models' characteristics. Furthermore, we transferred this knowledge to build and introduce the first Nav1.5 liability classification model in the field of computational toxicology, achieving an accuracy of Q2 = 86.7% for binary classification and multiclass classification accuracy of Q4 = 74.9%.

## 2. Related Work

Over the last two decades, scientists and researchers made significant use of several ML algorithms to build robust models to predict mainly hERG inhibition. In their review papers, Wang et al. [7] and Villoutreix et al. [8] summarized previous hERG liability ML models published in the field. Based on their findings, the most used machine learning algorithms in the early 2000s were Bayesian, Partial Least Squares(PLS), and Neural Network Inducers. However, during the current decade, scientists shifted towards the use of Random Forest (RF) and Support Vector Machine (SVM) due to their higher empirical performances in the field.

As of recent related works on hERG classification models, Czodrowski et al. [9] trained a RF model with descriptors calculated by Rdkit [10] on 3,721 compounds measured in a biding assay and 765 compounds measured in a functional assay from the ChEMBL database [11] showcasing a mean prediction accuracy of ~0.80. In a different research publication, Shen et al. [12] used SVM to build a predictive model combining both 4D-fingerprints and 2D and 3D molecular descriptors on a training set of 876 compounds from the PubChem BioAssay database [13]. Their model achieved an accuracy of 0.87 on an evaluation set of 456 compounds. Another interesting work using Bayesian classification models was published by Liu et al. [14]. The group gathered a dataset of 2,389 compounds from the FDA-approved drugs and from instances tested on hERG in the literature. Then, they trained their model using four molecular properties (MW, PSA, AlogP, and pKa_basic), as well as extended-connectivity fingerprints (ECFP_4). The development set was split into a training set of 2,389 instances and a test set of 255 compounds. The model achieved an accuracy of 0.91 on the test set and was further evaluated in another work by Doddareddy et al. [15] on an external set of 60 molecular compounds achieving an accuracy of only 0.58. In 2015, Braga et al. [16] deployed a model, named Pred-hERG, in a web-based platform to predict hERG liability. The model was trained on 5,984 compounds combining both Morgan fingerprints and Chemistry Development Kit descriptors using RDkit and PaDEL[17]. The model was reported to achieve a correct classification rate of 0.84. Whereas, the most recent hERG inhibition model published in 2019 by Kumar et al. [6] represents a consensus model following a one-vs-rest approach to build a pipeline of trained RF models to perform a multiclass classification at different inhibition levels of a given molecular compound. By far, Kumar's model was the most superior in terms of making use of data as well as performance. It employed 8705 claimed unique compounds from the ChEMBL database. Their methodology consisted of applying different sets of 2D descriptors to build each sub-model in the pipeline. The authors claimed achieving state-of-the-art performance with a Q2 = 0.92, sensitivity = 0.963, specificity = 0.786, and MCC = 0.786 evaluated on an external set of 499 compounds from PubChem and literature mining.

## 3. Methods

Figure 1 [18] represents a graphical representation of the hERG and Nav1.5 models building procedure abiding by the best practices of data science and QSAR methodology. Our methodology involves data gathering; curation, and normalization; feature generation and feature selection; splitting data into training, validation, and test sets; fine-tuning and building a set of models to pick the most performing model with the optimal configuration of hyperparameters; and finally evaluation of the best model on the test set.

### 3.1. Data Collection and Curation

Preparing data for both hERG and Nav1.5 is the main step in our machine learning models building pipeline. Artificial Intelligence (AI) predictive models can be very sensitive if wrong or biased data is used during learning. Hence, careful curation of all data entries was essential to have reliable models and results. Below is a description of the process we followed in data collection, curation, and organization while constructing the hERG and Nav1.5 models.

#### 3.1.1. hERG Data Collection and Curation

For hERG liability predictions, we gathered data from literature and other open bioactivity data sources. We started first by the dataset published by Kumar et al. [6], as one of the largest gathered and preprocessed hERG set publicly available to the research community. The set was derived from three data sources of molecular compounds [12], including ChEMBL v22, PubChem, and external instances from the literature. Data collected from ChEMBL v22 [19] resulted in 8705 unique molecular compounds. From PubChem BioAssay [11, 6, 20], 9,383 molecular records were retrieved under the gene id 3757 [21]. Kumar and his fellow researchers performed initial data curation on this set and reported 277 unique molecules.

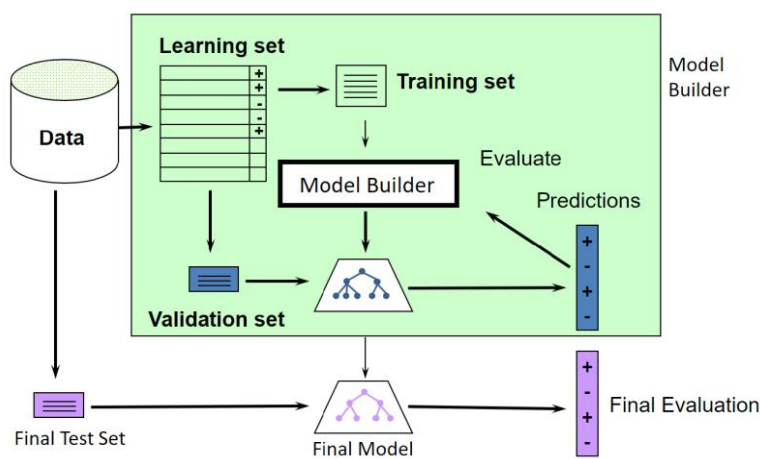

**Figure 1:** QSAR/Data Science best practices to perform best model selection and battle overfitting

From the literature mining they got a set of 561 compounds from the published research work by Li et al. [22]. Then, they performed normalization and redundancy reduction to report 222 unique molecular records.

We conducted further investigations and checks on the whole set of 9204 compounds and found many anomalies (detailed procedure to clean the set is explained in the Supplementary Online Material, section 1). The curation process led to a high-quality final set of 8879 entries, where 8380 unique compounds are retrieved from ChEMBL, and used solely for training, while the remaining 499 compounds gathered from both literature and PubChem were used for final evaluation only

*3.1.2.   Nav1.5 Data Collection and Curation*

To build a Nav1.5 model, we made use of two publicly available databases, namely the ChEMBL v25 database and the PubChem database. We retrieved 2051 molecular compounds from the ChEMBL v25 database [9, 23, 24] with Nav1.5 liability information. Similar to hERG data curation and standardization process, different potency records were reported by different laboratory assays. These data entries had records reporting $IC_{50}$, $K_i$, and $EC_{50}$ potency values. To keep data conformity, only records reporting $IC_{50}$ values were considered. For the rest of the redundant observations reporting different potency values, redundancy reduction procedure was conducted. The curation process was done as follows: First, for the same compound with multiple potencies, only the latest reported value in the last reference was kept. Second, if no latest reference is available, the mean value was computed and discarded the observations with large standard deviations. Finally, we normalized the $IC_{50}$ value for each unique compound by converting it to a negative logarithmic value, named the $PIC_{50}$. The curation process resulted in 1655 unique normalized records with reported $IC_{50}$ values through assays.

From the second publicly available dataset, namely, PubChem BioAssay [20], 2381 molecular records were retrieved under the gene id 6331 [25] (*SCN5A - sodium voltage-gated channel alpha subunit 5 for human*). Manual curation again of all the records was carried out. From the retrieved compounds, 721 molecules were discarded as they had no potencies reported, resulting in 1660 instances with potencies. To keep data conformity, only records reporting $IC_{50}$ values were considered, resulting in 1654 molecules with $IC_{50}$ potencies including redundant elements. Redundancy reduction was then conducted to filter out duplicate molecular compounds (detailed procedure is explained in the Supplementary Information, section 2). The final Nav1.5 development set consists of 1550 unique molecular compounds, denoted as Dev-Set-Nav in the rest of this manuscript, while the final evaluation set contained 173 instances, denoted as EV-Set-Nav and visualized in Figure S4-b in the Supplementary Information section.

*3.2. Machine Learning Algorithms*

In this work, we opted for Python as the current widely used computational and statistical programing language. To classify blockers from non-blockers, we adopted for state-of-the-art supervised machine learning algorithms used for classification. The three classifier algorithms we studied are:

- Deep Learning (DL/MLP) [26, 27, 28]
- Kernelized SVM (SVM) [29,30]
- Random Forest (RF) [31, 32]

Additionally, hERG liability predictions were analyzed using both regression models and pure classification models. For Nav1.5 data, PCA was used to reduce the high dimensional feature set of descriptors before building the machine models.

*3.2.1.   Principal Component Analysis (PCA)*

Data often lies on a low dimensional manifold embedded in a higher dimensional space. PCA [35] can help finding an approximation of the data in a target low dimensional space. The approach adopted by PCA is to find a new coordinate system in which the data points, that might be originally correlated, are linearly uncorrelated. After the linear transformation, dimensions with no or low variance can be ignored. In other words, we can apply a feature/dimension selection based on the principal components exhibiting higher eigen values. Given a dataset $X \in \mathbb{R}^{N \times D}$, where $N$ is the number of instances and *D* is the dimensionality of the features in the original set, applying PCA to reduce dimensionality from *D* to *K* dimensions is outlined in the mathematical notations as follows:

a. Center the data by deducting the mean (This can also be achieved via z-score normalization)
$$\widetilde{X} = X - 1_N(\frac{1}{N}1_N^T X) \quad (1)$$
where $\bar{x} = \frac{1}{N}1_N^T X$ is the mean value

b. Compute the covariance matrix
$$\Sigma = \frac{1}{N}\widetilde{X}^T \widetilde{X} \quad (2)$$

c. Compute the eigen decomposition of Σ
$$\Sigma = \Gamma^T \Lambda \Gamma \quad (3)$$

d. Pick the top-K eigenvectors $\Gamma_K$ corresponding to the K largest eigenvalues of the covariance matrix Σ

e. Project the data
$$Y = \widetilde{X} \Gamma_K \quad (4)$$

A main question to ask here is; How to choose K? The 90% rule is frequently used by practitioners in the field. We usually pick the first K variances/eigenvalues explaining 90% of the energy.

### 3.2.2. Deep Learning/MLP

The first supervised learning classifier investigated was a feed-forward network, also called a multi-layer perceptron (MLP) or artificial neural network [25, 26, 27]. The ultimate goal of using an MLP is to project input data points to a new dimensional space, also called latent space, using non-linear basis functions to allow for better probabilistic decision boundaries. This is achieved by fixing the number of basis functions in advance but allowing them to be adaptive via the use of parametric forms for those basis functions in which the parameter values are adapted/learned during training time [33]. All those functions are represented as a compute graph in a network, and their parameters are adapted via the use of a loss function that guides the training/update of all the parameters of the network during what is called back-propagation. For larger problems, neural networks may contain multiple layers, each with a predefined number of hidden neurons, stacked on top of each other to increase the capacity of the network. A multiple stacked layers network is referred to as a deep neural network or deep learning (DL) within the AI community. Figure 3 represents a network diagram of a two-layer neural network [33].

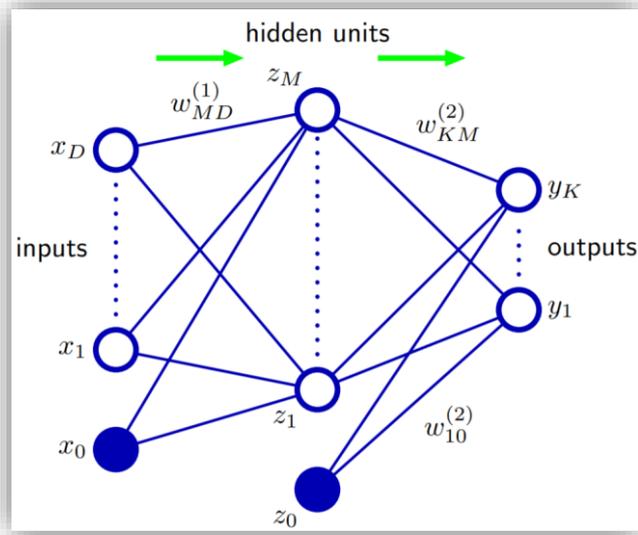

**Figure 3:** Two-layer neural network diagram. The input, hidden, and output variables are represented by nodes, and the weight parameters are represented by links between the nodes, in which the bias parameters are denoted by links coming from additional input and hidden variables $x_0$ and $z_0$. Arrows denote the direction of information flow through the network during forward propagation [30]

### 3.2.3. Kernelized SVMs

Support Vector Machine (SVM) is a well-known algorithm used for binary classification of data points. It is an algorithm categorized under the family of maximum margin classifiers. The main intuition behind the algorithm is that a wide margin around the dividing decision boundary makes the new samples more likely to fall on the correct side of the separating line. An important property of SVMs is that the determination of the model parameters corresponds to a convex optimization problem, and so any local solution is also a global optimum [25]. The objective function of SVMs is mainly solved via LaGrange multipliers [25] resulting in a dual function that can be re-written as a quadratic function which then can be solved with many efficient algorithms for quadratic programming problems. LIBSVM [31] is a widely used algorithm to solve such quadratic problems for SVMs and it is implemented in *sklearn.svm.SVC* [36] python library that we used in our modeling analysis. Additionally, since the data points may not be linearly separable (including noisy data) we opted for two solutions to elevate this problem: The use of the kernel trick [30] for non-linear decision boundaries, and the relaxation/punishment of the constraints by introducing slack variables symbolized as $\varepsilon_i$. The above explanation is formally summarized in the following constrained objective function that will be optimized using LaGrange multipliers:

$$\text{minimize} \quad f_0(\boldsymbol{w}, b, \boldsymbol{\varepsilon}) = \frac{1}{2}\boldsymbol{w}^T\boldsymbol{w} + C\sum_{i=1}^{N}\varepsilon_i \quad (5)$$
$$\text{subject to} \quad y_i(\boldsymbol{w}^T\boldsymbol{x_i} + b) - 1 + \varepsilon_i \geq 0$$
$$\varepsilon_i \geq 0$$

*3.2.4. Random Forest*

Random forest (DF) [31, 32] algorithm is the third investigated approach to classify blockers from non-blockers. It is an ensemble method of multiple decision trees used to make a classification decision based on the majority vote. Decision trees implement a greedy heuristic algorithm to build the tree using a top-down approach by picking the best splitting features with the best improvement based on a given impurity measure. An ensemble of decision trees usually suffers from the problem of high variance which may result in biased results.

To elevate this problem, we combined the decision forest with bagging (bootstrap aggregation) method to reduce the variance of the model. The main idea behind bagging is to create an ensemble of decision trees from randomly selected samples of data points from the training set with replacement. We made use of *sklearn.ensemble.RandomForestClassifier* [37] python library implementing a bootstrap aggregation RF.

*3.3. Feature Generation and Selection*

To perform statistical analysis and build our models, we needed numerical metrics for each molecular compound instance. The features used in this setting were called molecular descriptors. A molecular descriptor is the final result of a logical and mathematical procedure that transforms chemical information encoded within a symbolic representation of a molecule into a useful number or the result of some standardized experiment [38]. We made use of PaDEL-Descriptor v2.21, a tool implemented in Java by Yap Chun Wei [17, 39], to compute the molecular descriptors. The tool takes as an input a ".*smi*" file containing the smile format of the molecular compound and outputs a CSV file of the set of descriptors. It currently computes 1875 descriptors, including 1444 1D and 2D descriptors and 431 3D descriptors, and 12 types of fingerprints (a total of 16,092 bits).

Literature manuscripts reported that 2D descriptors provide better predictive results, and at the same time, require less computational resources compared to 3D descriptors [6, 40 - 42]. Therefore, we compiled only 2D descriptors using PaDEL-Descriptor v2.21 [17, 39] with a maximum runtime of $10^5$ milliseconds per molecule. The process resulted in 1444 2D descriptors. As the number of generated features was huge, feature selection and dimensionality reduction were required.

*3.3.1. Feature Selection for hERG Dataset*

As a first step of our feature selection procedure, we started by percent missing value analysis combined with information analysis of each single feature. This step reduced the feature space from 1444 to 1364. We followed this step by a pairwise correlation analysis supported by a predictive power strategy to decide on which feature to drop in case of collinearities of the features. The analysis reduced the set to 208 features using a correlation cut-off = 65%. To further reduce feature set, we applied LASSO as a context-dependent and an embedded dimensionality reduction technique. Splitting the development set into 80% training and 20% validation, we run a grid search to find the best regularization parameter $\lambda$. $\lambda=0.01$ was found to be the best hyperparameter. Applying LASSO on the whole development set of 8380 molecular compounds, the final feature space was reduced to 144 best features modelling the data.

*3.3.2. Feature Selection for Nav1.5 Dataset*

Through a percent missing value and an information analysis of each single feature, we reduced the feature space from 1444 to 551 (details in Supplementary Information section 3). The number of features was still high, therefore, a variance-based analysis was conducted as shown in figure 4. The graph gives a nice representation of the importance of features in our data based on their variances, ranging from $10^{-5}$ to $10^{21}$. The visual shows that an arbitrary cut-off at $10^{10}$ would lead to 7 features to be selected. However, this analysis was done on the original dataset and picking such cut-off, or even a bit lower for more features, may lead to wrong results. Also, applying a principal component analysis (PCA) directly on this data will result in biased and wrong analysis, as PCA will tend to focus more on the features with a high variance. The 27-order of magnitude difference of the variances is mainly due to the drastic different scales of the data features and the existence of outliers. Therefore, standardization and normalization of our feature values proves itself to be crucial. We used standard scaling, also called Z-score normalization, in our analysis to elevate this problem:

$$\text{Z-score} = \frac{x - \mu}{\sigma} \quad (6)$$

Where $x$ is the value to be normalized, $\mu$ is the empirical mean value of the column and $\sigma$ is the empirical standard deviation. The result of the normalization transformation of the data features is visualized in Figure S5 available in the Supplementary Information section. The figure shows how the variances of each of the 551 features are distributed after the standard normalization applied using *StandardScaler* [43] library implemented in python. The graph demonstrates that 7 features are of great importance in our dataset, as they exhibit the highest variance. However, since all the variance values fall between 0.95 and 1.05(tight range), a variance-based feature selection is hard to make, and an arbitrary cut-off might still provide wrong results. Also, some of the selected features could be highly correlated, which would include some bias to the trained models. The idea is then to decorrelate the system first, and then to base the selection of features on their energy.

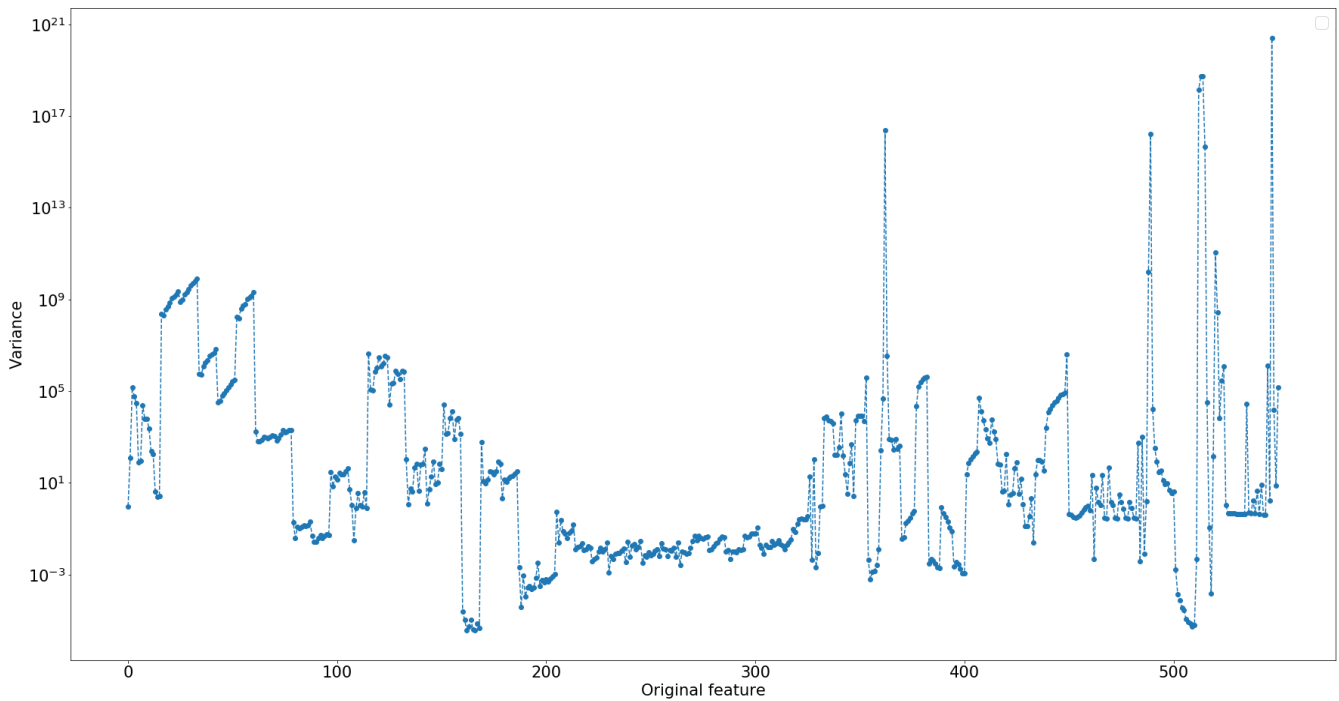

**Figure 4:** Dev-Set-Nav variance-based analysis of the 551 features. Very few features exhibit very high variance, and many are very low with a 27-order of magnitude variance difference

To achieve that, we conducted an unsupervised learning study called principal component analysis (PCA) on the entire normalized Dev-Set-Nav of 551 features. The results will be discussed in Nav1.5 result section.

*3.4. Training, Validation, and Test Set Splitting*

To abide by QSAR and data science best practices, we applied different strategies of model validation for each algorithm. For the DL classifier, we split the development set into 90% training and 10% validation sets as we wanted to use most of our data for training. However, for the kernelized SVMs and RF training, we used a technique named *k*-fold cross-validation via scikit-learn library in Python [44].

In *k*-fold cross-validation (CV), sometimes called rotation estimation, the dataset $D$ is randomly split into $k$ mutually exclusive subsets (the folds) $D_1, D_2, ..., D_k$ of approximately equal size. The inducer is trained and validated k times; each time $t \in \{1,2,....,k\}$ it is trained on $D \backslash D_t$ and tested on $D_t$. After a grid-based search for the optimal number of folds, we opted for 10-fold cross-validation for our hyperparameter optimization of the sub-models, as being the one giving the best CV estimate of accuracy for both SVM and RF. The cross-validation estimate of accuracy is the overall number of correct classifications, divided by the number of instances in the dataset. It may also be seen as the mean accuracy of all the folds accuracies in case of stratified split. Formally, let $D_{(i)}$ be the validation set that includes instance $x_i = (v_i, y_i)$, then the cross-validation estimate of accuracy [45] is:

$$AC_{cv} = \frac{1}{n} \sum_{(v_i,y_i) \in D} \delta\left(I(D \backslash D_i, v_i), y_i\right) \quad (7)$$

Applying just a plain CV may lead to biased accuracy estimates of the validation set. The problem is frequently present when the data is not equally distributed among classes, which causes a random selection of *k*-folds to have a dominance of instances, often only instances, of the majority classes. Therefore, to elevate this issue, we used a technique called stratification for both normal training/validation split, case of MLP, and *k*-fold CV, case of SVM and RF training. A stratified split is a split that maintains a similar percentage distribution of classes in the original set.

*3.5. Performance Evaluation Metrics*

As we are following the approach of one-vs-rest methodology for multiclass classification, we used a set of six measurements for binary classification during our modeling analysis evaluation to assess the quality of trained models:

$$AC = AC_{bin} = \frac{TP+TN}{TP+FN+TN+FP} \quad (8)$$

$$SN = \frac{TP}{TP+FN} \quad (9)$$

$$SP = \frac{TN}{TN+FP} \quad (10)$$

$$F1 = 2 * \frac{SN*PR}{SN+PR} \quad (11)$$

$$CCR = \frac{SN+SP}{2} \quad (12)$$

$$MCC = \frac{(TP*TN)-(FP*FN)}{\sqrt{(TP+FN)*(TP+FP)*(TN+FN)*(TN+FP)}} \quad (13)$$

Where TP stands for true positive and represents the number of correctly predicted blockers, TN stands for true negative and represents the correctly predicted non-blockers, FP stands for false positive and represents the number of non-blockers predicted as blockers, and FN stands for true negative and counts the number of blockers predicted as non-blockers.

Here, we want to stress on the F1 measurement, also called the F-score. It considers the sensitivity and the precision in its formula measuring the harmony of both metrics. F1 score, as a comprehensive statistical parameter, is considered to be a better evaluation criterion than accuracy when dealing with imbalanced datasets [40], which is our case.

The above metrics are used in the context of binary classification. For multiclass classification measures, we provide a multiclass accuracy denoted as $AC_{mul}$ and formally computed as follows:

$$AC_{mul} = \frac{1}{n} \sum_{(v_i,y_i) \in T} \delta\left(I(v_i), y_i\right) \qquad (14)$$

Where $T$ is the set of instances used for evaluation containing observations each one denoted by the pair $x_i = (v_i, y_i)$. Whereas, $I$ function represents the inducer or the predictor, and $\delta$ is the indicator function returning 1 if the predicted class is matching the ground truth and 0 otherwise.

Whereas, for regression we considered two measurements, the Sum of Squared Error and the R-Squared metric:

$$MSE = \frac{1}{n} \sum_{i=1}^{n}(y_i - \hat{y}_i)^2 \qquad (15)$$

$$R^2 = 1 - \frac{SS_{RES}}{SS_{TOT}} = 1 - \frac{\sum_{i=1}^{n}(y_i - \hat{y}_i)^2}{\sum_{i=1}^{n}(y_i - \bar{y})^2} \qquad (16)$$

### 3.6. Data labeling and Sampling Techniques for Classification

With multiclass classification algorithms, QSAR models were trained on datasets of molecular compounds labeled as follows: A compound with $PIC_{50} \geq 6$ is labeled as "strong blocker", a compound with $6 > PIC_{50} \geq 5$ is labeled as "moderate blocker", a compound with $5 > PIC_{50} \geq 4.5$ is labeled as "weak blocker", and a compound with $4.5 > PIC_{50}$ is labeled as "non-blocker".

Following a one-vs-rest approach to build a multiclass classifier, QSAR models were built and analyzed on 3 differently labeled development sets on each of hERG and Nav1.5 curated and normalized datasets. From each curated development set, we derived three labeled datasets by applying each time a different threshold based on the $IC_{50}$ values of 1, 10, and 30μM, corresponding to the $PIC_{50}$ values of 6, 5, and 4.5, which corresponds eventually to the level of potency blockage intensities of strong, moderate, and weak blockers respectively.

For hERG, each time we applied a different threshold, we obtained a different class distribution of hERG blockers(blk) and non-blockers(nblk). For a threshold of $PIC_{50} = 6$, the class distribution was (blk: 1596 vs nblk: 6784). For a threshold of $PIC_{50} = 5$, the class distribution was (blk: 5116 vs nblk: 3264). For a threshold of $PIC_{50} = 4.5$, the class distribution was (blk: 7003 vs nblk: 1377). We observed the same case for Nav1.5 after applying a different threshold each time. We obtained imbalanced class distribution of Nav1.5 blockers(blk) and non-blockers(nblk). For a threshold of $PIC_{50} = 6$, the class distribution was (blk: 160 vs nblk: 1390). For a threshold of $PIC_{50} = 5$, the class distribution was (blk: 1018 vs nblk: 532). For a threshold of $PIC_{50} = 4.5$, the class distribution was (blk: 1442 vs nblk: 108).

To distinguish which development set was used for building a particular model, the $PIC_{50}$ potency value applied for compounds tagging on the dataset was attached to the ML algorithm used (ex. '6RF' means that the random forest is trained on a dataset labeled according to a cut-off of $PIC_{50} = 6$). For the two thresholds of 6 and 4.5, the class distributions were very imbalanced. Such data distributions make learning algorithms to often be biased towards the majority class as the algorithms main focus is trying to reduce the error rate without taking the data distribution into consideration. There exist four main ways to deal with class imbalances: re-sampling, re-weighing, adjusting the probabilistic estimate, one-class learning. In our research work, we opted for re-sampling methodology to deal with this issue as a simple approach of biasing the generalization process. we wanted to analyze the results of both a balanced dataset and an imbalanced one. We tried two different approaches for this analysis: over-sampling and under-sampling. For over-sampling, we used the Synthetic Minority Over-sampling Technique (SMOTE). SMOTE generates synthetic examples in a less application-specific manner, by operating in "feature space" rather than "data space" [46]. The minority class is over-sampled by taking each minority class sample and introducing synthetic examples along the line segments joining any/all of the $k$ minority class nearest neighbors. Depending on the amount of over-sampling required, neighbors from the $k$ nearest neighbors are randomly chosen. Chawla et al. explained the synthetic generation of samples as follows: First, we take the difference between the feature vector (sample) under consideration and its nearest neighbor. Second, we multiply this difference by a random number between 0 and 1 and we add it to the feature vector under consideration. This causes the selection of a random point along the line segment between two specific features. This introduced algorithm effectively forces the decision region of the minority class to become more general [46]. The approach was applied via the *imblearn.over_sampling* [47] library in Python implementing the explained SMOTE algorithm. Concerning under-sampling, we used NearMiss algorithm, a negative example selection method [48], implemented in the Python library *imblearn.over_sampling* [49].

### 4. Results and Discussion

All computations reported in this manuscript were conducted on a personal workstation. The machine system memory size was 16Gib with an Intel® Core™ i7-8750H 8[th] generation CPU with 12 logical processors and maximum speed of 3.5 GHz. The machine contains also a NVIDIA GeForce GTX 1050 Ti with a dedicated memory of 4Gib. The GPU was used for deep learning hyperparameter tuning.

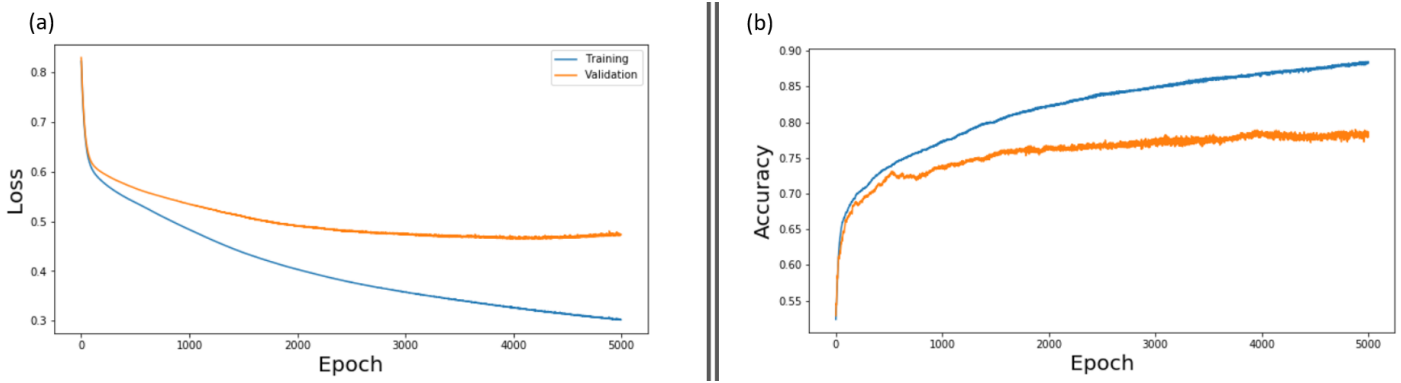

**Figure 5:** Training results of the best 2-layer network with 40N(HL). (a) represents the loss change with respect to the number of epochs of both training and validation sets. (b) represents the accuracy change with respect to the number of epochs of both training and validation sets. The best hyperparameter configuration is: L2-norm weight decay, 512 batch size, ReLu as activation function, and Xavier initialization

*4.1. hERG ML Models – Results and Discussion*

For hERG multiclass predictions, we opted for the one-vs-rest approach. To pick the best performing ML algorithm to use in our final deployed sub-models, we made our first algorithm selection analysis using one labeled dataset with a fairly balanced class distribution. The threshold of choice was $PIC_{50} = 5$ with a class distribution of 5117 hERG blockers and 3265 non-blockers.

*4.1.1. Machine Learning Algorithms Analysis applied on Literature-Based Feature Selection*

For literature-based features, we gathered all the collections of best features compiled in [6] and combined them in one unique set of best features to be used for development. This set of features were selected based on a greedy algorithm named Best-First Feature Selection method. We ended up with 191 unique literature-mined features, selected from the 1444 2D descriptors. We used the final set for training all the sub-models of our *ToxTree- hERG Classifier*. This set was used to analyze and then select the best performing classifier algorithm to choose among the MLP, SVM, and RF.

*4.1.1.1. Deep Learning / MLP*

In our analysis, we used a grid-based search for hyperparameter tuning in order to find the best hyperparameter configuration of our simple artificial neural network to classify correctly the hERG blockers and non-blockers.

The search space included one single network architecture of 2 layers (1 hidden layer of 40 neurons and an output layer of two neurons), fixed learning rate of $10^{-3}$, fixed L2-norm weight decay, 2 types of activation functions (sigmoid and ReLu), 50% dropout rate [50] or not, batch normalization [51] or not, and 2 batch sizes: 256 and 512. The search space resulted in 16 different models, all initialized using Kaiming(for ReLu) or Xavier(for Sigmoid) initialization [52]. To select the optimal model, the best hyperparameters of the architecture were saved if the loss on the validation set scored lower than the last saved best model after a complete epoch.

Figure 5 represents the loss and accuracy results of the best 2-layer network found during our parameter search, after 5000 epochs. The curves show that the model converged around 4000 epochs, and the generalization gap started increasing afterwards (figure 5-a). The best model hyperparameters achieved a validation accuracy of 76.1%.

*4.1.1.2. Kernelized SVMs*

We applied again grid-based search to identify the best parameters of the optimal kernelized SVM for our given problem using a 10-fold stratified cross-validation. The parameters that were used in the grid search included 4 kernel functions (linear, polynomial, sigmoid, and radial basis function kernel). In the case of a polynomial kernel, 9 degrees were investigated (from 2 to 10). For constraints relaxation, 10 values were evaluated of the penalty constant C, with 5 values less or equal than 1 (0.1,0.2,0.5,0.8, 1) denoting a soft margin SVM and 5 greater than 1 (3, 5, 10, 50, 100) denoting hard margin SVM. The Search space resulted in 120 different models built with 4 types of kernels. The best performing model achieved an $Ac_{cv}$ of 75% with a polynomial kernel of degree 2 and a penalty constant of 0.2.

*4.1.1.3. Random Forest*

To evaluate the performance of Random Forest on our given training dataset, we applied a grid-based hyperparameter search combined with a 10-fold stratified cross-validation. The search space included 100 different configurations to evaluate. we achieved a best $Ac_{cv}$ of 81.1% with an ensemble of 100 decision trees.

### 4.1.2. Best Performing ML Algorithm Based on Literature-Mined Features

Among the three evaluated ML algorithms (sections 4.1.1.1, 4.1.1.2, and 4.1.1.3), RF represented the best performing model with a validation accuracy of 81.1% compared to the kernelized SVM and the MLP with accuracies of 75% and 76.1% respectively. Therefore, to build our ToxTree-hERG Classifier model, RF with bootstrapping was our algorithm of choice to build our one-vs-rest sub-models at the three potency intensity levels.

### 4.1.3. Model Building and Evaluation on External Test Sets using Literature-Mined Feature Set

We conducted a grid-based search on a set of 11 different values starting at 10 and incrementing by 10 for each next value to define the optimal number of estimators (decision trees) per RF classifier for each of the three separately labeled datasets. The hyperparameter search was applied using 10-fold stratified CV technique. For each $PIC_{50}$ threshold, we compared the results of balanced datasets with imbalanced ones. Applying the two previously explained sampling methods (over/under-sampling), we reported the results in Table 1. The statistics include the cross validation estimate of accuracy ($Ac_{cv}$), defined in eq. 7, as well as the cross validation estimate of F1-score ($F1_{cv}$), computed similarly based on single folds F1-scores.

Table 1 shows the results of the best performing ensemble of decision trees for each labeled development set. For the threshold $PIC_{50} = 6$, we picked 6rf-ovrs model distinguishing 'strong hERG blockers' from non-blockers as being the best performing one, with an $AC_{cv}$ of 93.3%, 80-ensemble of estimators, and a maximum tree depth of 57. For the threshold of $PIC_{50} = 4.5$ tagging 'weak-blockers' and 'non-blockers', two models were reporting high $Ac_{cv}$ and comparable $F1_{cv}$. They were both selected to construct the final Model. Inducers 4o5rf and 4o5rf-ovrs reported an $Ac_{cv}$ of 88.7% and 94.8%, $F1_{cv}$ of 93.5% and 94.7%, number of estimators per RF of 80 and 80, and a maximum depth of the trees of 51 and 47 respectively. Whereas, we kept only the best performing classifier, 5rf-ovrs, trained on an over-sampled development set with synthetic samples for the $PIC_{50} = 5$ threshold labeled dataset to classify 'moderate blockers.' 5rf-ovrs achieved a $AC_{cv}$ of 85.4% with a maximum tree depth of 35 and an ensemble of 110 estimators.

The best hyperparameters along with best performing sampling techniques were used to build our final models trained on the entire 8380 molecular compounds for hERG liability predictions. Table 2 shows that our individual built models exhibit higher performances than all the models presented by Kumar et al. [6]. Our models outperformed all the models with respect to the overall accuracies and the F1-scores on the untouched test sets, EV-1 and EV-2. Whereas the specificity values are comparable.

**Table 1:** The cross-validation estimate of accuracy rate ($AC_{cv}$) and F1-score ($F1_{cv}$) of the best performing RF model in each of the 3 differently labeled training datasets based on the three $PIC_{50}$ cut-offs and different sampling strategies. Over/under-sampling analysis was carried out at all of the three potency thresholds as training data was imbalanced. $AC_{cv}$ and $F1_{cv}$ measurements were calculated following a 10-fold-stratified-cross-validation applied on the training dataset, i.e. computed on validation folds. The models with the best CV estimate of accuracy (eq. 7) rates are mentioned in bold. Similarly, we computed the CV estimate of F1-score and reported the best inducer with its corresponding number of estimators and max depth in the RF. Grid-based search for the optimal number of estimators was conducted on 11 different values starting at 10 and incrementing by 10 for each next value.

| $PIC_{50}$ Threshold | Model short name | Sampling strategy | Development class distribution | | $AC_{cv}$ | $F1_{cv}$ | Num of estimators | Max depth |
|---|---|---|---|---|---|---|---|---|
| | | | blk | nblk | | | | |
| 4.5 | **4o5rf** | **Original** | **7003** | **1377** | **88.7** | **93.5** | **80** | **51** |
| | **4o5rf-ovrs** | **Over-sampling** | **7003** | **7003** | **94.8** | **94.7** | **80** | **47** |
| | 4o5rf-unds | Under-sampling | 1377 | 1377 | 80.8 | 80.6 | 90 | 38 |
| 5 | 5rf | Original | 5116 | 3264 | 81.1 | 84.8 | 100 | 44 |
| | **5rf-ovrs** | **Over-sampling** | **5116** | **5116** | **85.4** | **85.2** | **110** | **35** |
| | 5rf-unds | Under-sampling | 3264 | 3264 | 80.4 | 80.4 | 100 | 51 |
| 6 | 6rf | Original | 1596 | 6784 | 87.3 | 57.1 | 90 | 43 |
| | **6rf-ovrs** | **Over-sampling** | **6784** | **6784** | **93.3** | **93.2** | **80** | **57** |
| | 6rf-unds | Under-sampling | 1596 | 1596 | 76.1 | 75.9 | 110 | 36 |

**Table 2:** Prediction performance of individual best performing models on the test sets EV-1 and EV-2 compared with the state-of-the-art sub-models of the consensus model introduced by Kumar et al. in [6]

| Test set | $PIC_{50}$ Threshold | Training dataset | Model | AC | SN | SP | F1 | TP | FN | TN | FP |
|---|---|---|---|---|---|---|---|---|---|---|---|
| EV-1 | 4.5 | Original/Imbalanced | **4o5rf** | 88 | 98 | 58 | **93** | 166 | 4 | 30 | 22 |
| | | | 4o5_RFfwd | 86 | 99 | 42 | 91 | 169 | 1 | 22 | 30 |
| | | Balanced | **4o5rf-ovrs** | 89 | 95 | 67 | **93** | 162 | 8 | 35 | 17 |
| | | | 4o5_s1_RFfwd | 82 | 88 | 62 | 88 | 149 | 21 | 32 | 20 |
| | | Combined | **Combined_4o5rf** | 90 | 98 | 64 | **94** | 167 | 3 | 33 | 19 |
| | | | 4o5Combined | 87 | 98 | 52 | 91 | 166 | 4 | 27 | 30 |
| | 5 | Original/Balanced | **5rf-ovrs** | 82 | 85 | 77 | **86** | 119 | 21 | 63 | 19 |
| | | | 5RFfwd | 77 | 72 | 84 | 79 | 101 | 39 | 69 | 13 |
| | 6 | Original/Imbalanced | 6RFfwd | 73 | 20 | 97 | 32 | 14 | 55 | 149 | 4 |
| | | Balanced | **6rf-ovrs** | 83 | 57 | 95 | **67** | 39 | 30 | 145 | 08 |
| EV-2 | 4.5 | Original/Imbalanced | **4o5rf** | 96 | 99 | 83 | **97** | 211 | 1 | 54 | 11 |
| | | | 4o5_RFfwd | 93 | 100 | 72 | 95 | 206 | 0 | 51 | 20 |
| | | Balanced | **4o5rf-ovrs** | 97 | 99 | 91 | **98** | 209 | 3 | 59 | 6 |
| | | | 4o5_s1_RFfwd | 82 | 80 | 90 | 87 | 164 | 42 | 64 | 7 |
| | | Combined | **Combined_4o5rf** | 96 | 99 | 88 | **98** | 210 | 2 | 56 | 8 |
| | | | 4o5Combined | 94 | 98 | 85 | 96 | 201 | 5 | 60 | 11 |
| | 5 | Original/Balanced | **5rf-ovrs** | 89 | 93 | 86 | **89** | 125 | 10 | 122 | 20 |
| | | | 5RFfwd | 77 | 67 | 87 | 74 | 90 | 45 | 123 | 19 |
| | 6 | Original/Imbalanced | 6RFfwd | 93 | 50 | 100 | 67 | 19 | 19 | 239 | 0 |
| | | Balanced | **6rf-ovrs** | 95 | 66 | 100 | **79** | 25 | 13 | 239 | 0 |

### 4.1.4. Comparison Between Literature-Mined and L1-Based Feature Selection Methods

Before applying the L1-based sparse feature selection method, LASSO, we applied a set of feature reduction techniques. Using percent missing value analysis, we first reduced the number of features from 1444 to 1364. Second, we applied a pairwise correlation analysis supported by predictive power analysis to guide the filtering out of most relevant features. This analysis reduced the feature space to 213. Finally, for a more context-dependent feature selection analysis, the embedded dimensionality reduction technique, LASSO, was applied. Splitting data to training and validation sets (80:20), the best MSE loss was achieved on the validation set at the regularization parameter alpha = 0.01. This reduced the feature space to 152.

Comparing the classification performance of each best feature selection methodology (Table 3) shows that the literature retrieved set of features (Best-First) outperforms the L1-based sparse feature selection method (Pairwise correlation + LASSO), especially in terms of specificity and F1-score. In our modeling, we want to achieve a good harmony between sensitivity and specificity. We don't want to favour only true positives or true negatives. Hence, literature-based selected features represent the final set of choice in building *ToxTree-hERG Classifier*.

**Table 3:** Prediction performance of individual best performing sub-models on the test sets EV-1 and EV-2. The table compares RF models trained on the same labeled datasets but using different set of features, either the literature retrieved set of features (following Best-First greedy algorithm), or L1-based sparse selected set of features (following a pairwise correlation feature reduction followed by LASSO).

| Test sets | $PIC_{50}$ Threshold | Training dataset | Feature selection Procedure | Model | AC | SN | SP | F1 | TP | FN | TN | FP |
|---|---|---|---|---|---|---|---|---|---|---|---|---|
| EV-1 | 4.5 | Original/Imbalanced | **Best-First** | **4o5rf** | **88** | **98** | **58** | **93** | **166** | **4** | **30** | **22** |
| | | | Pairwise correlation + LASSO | 4o5rf_Lasso | 84 | 98 | 38 | 90 | 166 | 4 | 20 | 32 |
| | | Balanced/over-sampling | **Best-First** | **4o5rf-ovrs** | **89** | **95** | **67** | **93** | **162** | **8** | **35** | **17** |
| | | | Pairwise correlation + LASSO | 4o5rf_ovrs_Lasso | 86 | 96 | 54 | 91 | 164 | 6 | 28 | 24 |
| EV-2 | 4.5 | Original/Imbalanced | **Best-First** | **4o5rf** | **96** | **99** | **83** | **97** | **211** | **1** | **54** | **11** |
| | | | Pairwise correlation + LASSO | 4o5rf_Lasso | 95 | 100 | 78 | 97 | 212 | 0 | 51 | 14 |
| | | Balanced/over-sampling | **Best-First** | **4o5rf-ovrs** | **97** | **99** | **91** | **98** | **209** | **3** | **59** | **6** |
| | | | Pairwise correlation + LASSO | 4o5rf_ovrs_Lasso | 96 | 99 | 86 | 98 | 211 | 1 | 56 | 9 |

*4.1.5. ToxTree-hERG Regressor*

Given the large dataset (8380 molecular compounds) with numerical values as a target, we have also built a regression model predicting the $PIC_{50}$ value. The classification performance of the regression model, *ToxTree-hERG Regressor*, was then compared to the one of multiclass inducer, *ToxTree-hERG Classifier*, on the same test set (section 4.1.7). The regression model was built using a Random Forest Regressor. This choice was supported by the fact that Random Forests are very powerful in depicting non-linear correlations between features; hence, further performing a feature selection step following another embedded greedy heuristic algorithm within the final regressor model. The best model achieved an *MSE* = 0.59 on the evaluation set of N=499.

Figure 6 represents the performance of the regressor in predicting the $PIC_{50}$ potency values of a molecular compound with respect to the ground truth values. The model is denoted as *ToxTree-hERG Regrossor* in the rest of the paper and achieves a coefficient of determination $r^2 = 0.67$ on the test set.

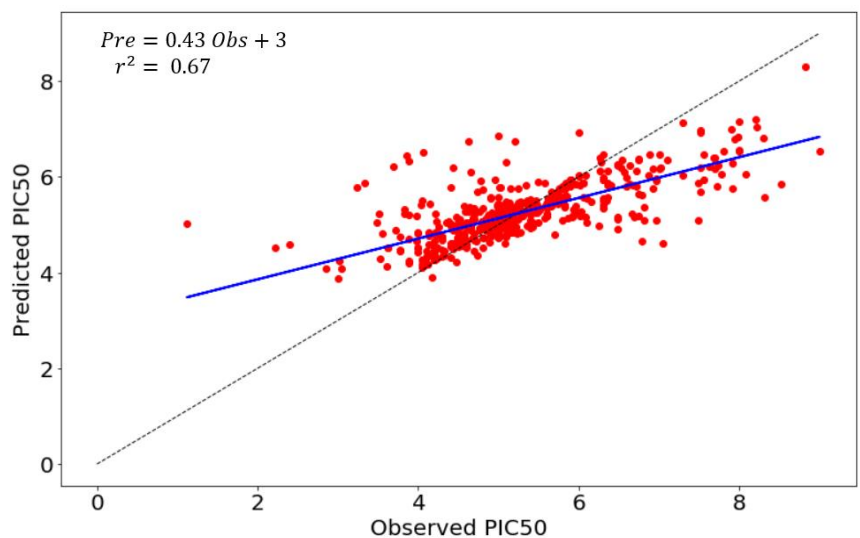

**Figure 6:** Regression scatter plot of the Predicted vs. Observed $PIC_{50}$ of our ToxTree-hERG Regressor model on the test set (N=499) combining both EV-1 and EV-2

*4.1.6. ToxTree-hERG Classifier*

As many reports and manuscripts [6, 53 - 55] reported that consensus QSAR models perform better than individual models, we combined 4o5rf and 4o5rf-ovrs in a single model named 'combined_4o5rf'. The decision making of the combined_4o5rf model is done as follows: First, if both 4o5rf and 4o5rf-ovrs predictions match, the new compound is assigned the same class. Second, if both classes differ, we pick the class inferred by the model reporting the highest probability. Finally, if both classes and their probabilities are similar, the decision is reported as inconclusive. The consensus/combined model also shows slightly higher performance than Kumar et al. combined model. To have a final multiclass classifier, we organized the one-vs-all hERG predictive models in a form of a binary tree to filter hERG blockers from non-blockers at the three potency thresholds. The models are applied sequentially in a pipeline. The final model (visualized in Figure 7) is denoted as ToxTree-hERG.

To briefly summarize ToxTree-hERG Classifier prediction process of a new molecular compound with unknow potency, we introduce the following steps: Given pre-computed 2D descriptors of a new molecular compound, we first select the 191 features used by our inducer system. Second, we normalize the data with the learned empirical mean and standard deviation. Third, we run 6rf-ovrs model to predict if it is a 'strong blocker'. If it is a non-blocker, we run 5rf-ovrs model to evaluate if it is a 'moderate blocker'. If it is a non-blocker, we run the combined_4o5rf consensus model to evaluate if it is a 'weak blocker', a 'non-blocker', or 'inconclusive'.

*4.1.7. ToxTree–hERG Benchmarking Evaluation*

In order to benchmark the performance of our ToxTree-hERG classifier and regressor models with existing hERG liability prediction tools in the market, a set of selected tools were evaluated on the combined test set of both EV-1 and EV-2, resulting in N=499 molecular compounds to be assessed.

Concerning the set of tools to be analyzed, we selected 9. Four of which are web-based servers, namely admetSAR [56], CoFFer [57], pkCSM [58], and Pred-herg [59]. Whereas, the remaining tools are standalone software, namely PaDEL-DDPredictor v2.1 [60], Schrödinger/QikProp [61], Simulation plus/ADMET Predictor [62], StarDrop v6.4 [63], and Consensus [6] model. The binary evaluation was not conducted inhouse by running all of those tools on the test set, but retrieved from Kumar et al. manuscript reporting the statistics. Some tools report/predict only the potency values.

Therefore, and for consistency of the results, the values were all converted to binary classes with a threshold of less than 30μM being a blocker and above as a non-blocker.

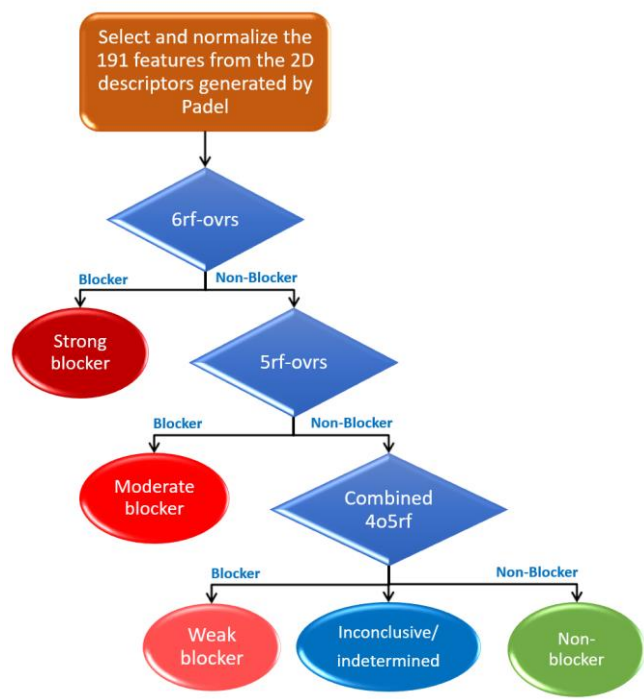

**Figure 7:** ToxTree-hERG Classifier conceptual visualization inference pathway

In Table 4, we notice that ToxTree-hERG Classifier is outperforming all the existing tools, including ToxTree-hERG Regressor, achieving a binary accuracy of Q2 = 93.2%, with an improvement of 1.2% to 53.1%. We also notice that ToxTree-hERG Regressor shows a high sensitivity compared to all the other models, which is mainly due to the high imbalanced training set for this binary task. This problem is elevated in ToxTree-hERG Classifier through the re-sampling technique (SMOTE). Another important mechanism allowing ToxTree-hERG to show good results is the architecture of the model that filters out the potency level of a molecular compound via multiple sub-models trained at different cut-offs. Though the dataset is still the same, each sub-model is trained on a different distribution of data points depending on the cut-off used.

Since we used a one-vs-all approach to discriminate hERG blockers from non-blockers, we can also evaluate the multiclass classification accuracy of our model. Figure 8 represents a multiclass confusion matrix at the three potency intensity levels. Using eq. 14, we computed the multiclass classification accuracy of ToxTree-hERG achieving 74.5% on the test sets of EV-1 and EV-2 both combined(N=499). The confusion matrix shows the power of the model in predicting 'moderate blockers' with an accuracy performance of 83.3% while predicting 'strong blockers' with a lower accuracy of 59.8%. We can also notice that a significant percentage of wrong classifications (purple cells) always happen in the neighboring classes. This is expected as we may always have data points very close to the decision boundaries of each sub-inducer in our model. As we are running the models sequentially, some of the molecular compounds having potency values less than the threshold or higher, but at the same time very close to it gets assigned to the previous or next class. From table 2 we can also infer that the Consensus model [6] achieves a 'strong blocker' classification accuracy of only 30.8%; while, we are achieving a 2x higher performance at this class. We again show how ToxTree-hERG Classifier outperforms the Consensus model in predicting strong blockers.

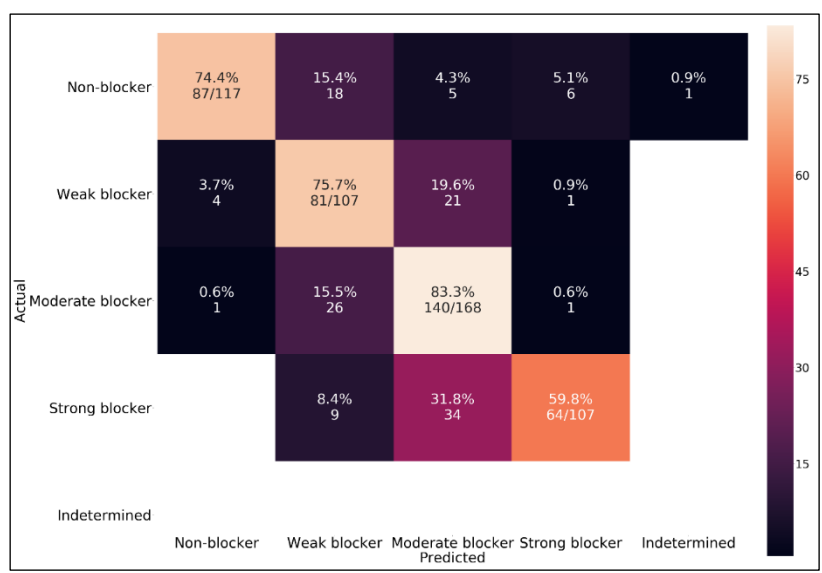

**Figure 8:** Multiclass confusion matrix of ToxTree-hERG inducer on the combined evaluation sets of EV-1 and EV-2

**Table 4:** Comparing prediction performance of ToxTree-hERG with the existing models and tools on a combined test set of both EV-1 and EV-2

| Inducer | AC | CCR | MCC | SN | SP |
|---|---|---|---|---|---|
| admetSAR | 68.1 | 57.2 | 27.9 | 78.7 | 35.8 |
| CoFFer | 40.1 | 51.8 | 20.6 | 28.7 | 74.8 |
| PaDEL-DDPredictor | 45.8 | 51.1 | 21.6 | 41.1 | 61.1 |
| pkCSM | 76.6 | 62.6 | 36.8 | 90.2 | 35 |
| Pred-herg | 70.5 | 77.2 | 49.8 | 64.1 | **90.2** |
| Schrödinger/QikProp | 75.1 | 58.8 | 30.3 | 90.5 | 27.1 |
| Simulation plus/ADMET Predictor | 74.3 | 64.4 | 38.3 | 84 | 44.7 |
| StarDrop | 75.8 | 57.4 | 28.1 | 93.6 | 21.1 |
| Consensus | 92 | **87.6** | 78.6 | 96.3 | 78.6 |
| ToxTree-hERG Regressor | 86.8 | 72.4 | 60.6 | **99.5** | 45.3 |
| ToxTree-hERG Classifier | **93.2** | 86.8 | **80.3** | 98.7 | 75 |

### 4.2. Nav1.5 ML Models – Results and Discussion
#### 4.2.1. Feature Multicollinearity Elimination

To reduce multicollinearity of the features, PCA was applied on the entire normalized Dev-Set-Nav of 551 selected features using *sklearn.decomposition.PCA* [64] python library. The library implements Singular Value Decomposition (SVD) [65] algorithm for matrix factorization. Figure S6 in the Supplementary Information section represents a visualization of the 551 principal components (PCs) with respect to their eigenvalues.

The graph shows that very few principal components are of great importance than others. However, as the visualization of the 551 PCs is condensed and hard to analyze, we chose to display the first 40 PCs with respect to their eigenvalues for better analysis, as shown in Figure 9. The blue curve represents the strength (explained variance) of each PC in the projected space. While the orange curve represents the cumulative explained variance, which grows as we increase the number of PCs. The green horizontal line represents the cut-off (90%) we chose to decide on the number of principal components space to which the data will be projected.

The intersection, the black star, shows that 90% of the energy is preserved at 22 principal components. Using eq. 4, we perform a linear transformation of our data were $\Gamma_K$ represents the learned matrix of the first k=22 eigenvectors and $\tilde{X}$ is the z-score normalized Dev-Set-Nav dataset of 551

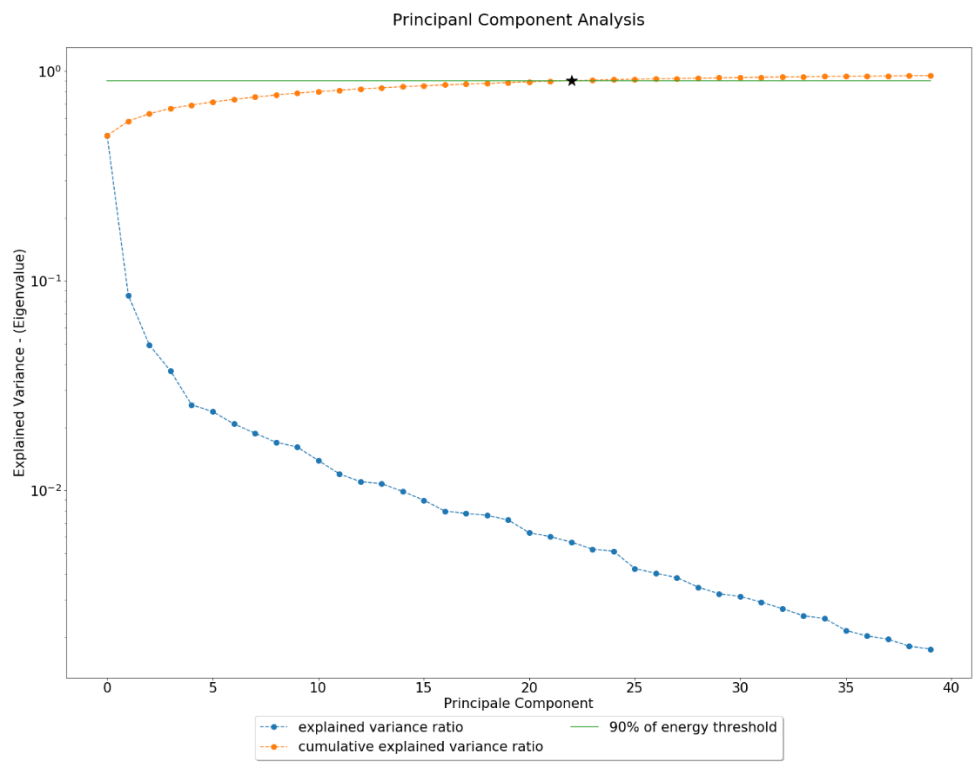

**Figure 9:** Principal component analysis(PCA) of the Dev-Set-Nav dataset 551 normalized features. The graph represents a visualization of the first 40 principal components (PC), with respect to their eigenvalues, as the visualization of the 551 PCs will not be clear. The blue curve represents the strength (explained variance) of each PC in the projected space. While the orange curve represents the cumulative explained variance, which grows as we increase the number of PCs. The green horizontal line represents the cut-off (90%) we chose to decide on the number of principal components space to which the data will be projected. The intersection, the black star, shows that 90% of the energy is preserved at 22 principal components

features. The new dimensionality of the reduced development set is now of 22 dimensions. The new reduced development set is denoted as Dev-Set-Nav-Red in the manuscript.

### 4.2.2. *Model building*

As our ultimate goal from this research is to build a multiclass classifier (non-blocker, weak-blocker, moderate-blocker, and strong-blocker) for Nav1.5 liability predictions, we applied three of the state-of-the-art machine learning algorithms. We investigated the performance of deep learning, random forest, and a one-vs-rest approach of kernelized SVM models. All models were trained on the normalized Dev-Set-Nav reduced to 22 features/PCs. For best model selection and hyperparameter tuning, a stratified split of 90% training and 10% validation was used to analyze deep learning models, and 10-fold stratified CV was applied in case of RF and SVM.

After a multiclass tagging and splitting of the Dev-Set-Nav-Red, SMOTE re-sampling strategy was used to elevate the problem of imbalanced training datasets and biasing the generalization process in the case of minority classes.

#### 4.2.2.1. *MLP Analysis*

For hyperparameters tuning, we applied grid-based search technique on 4 parameters (Activation Function, Dropout Rate, Batch Normalization, and Batch Size) where each one takes two different options. This resulted in 16 different configurations to evaluate on a 3-layer MLP of 40 hidden neurons in the first layer, 20 hidden neurons in the second, and 4 output neurons. The training of all models was conducted following Kaiming initialization in case of ReLu activation function or Xavier initialization in case of Sigmoid, a learning rate of $10^{-3}$, L2-norm weight decay, and optimized via Adam [66] optimizer. To pick the optimal model, the best hyperparameters of the architecture were saved if the loss on the validation set scored lower than the last saved best model after a complete epoch.

Table 5 shows the results of the optimal hyperparameters configuration of each of the 16 trained deep learning models after 4000 epochs. After labeling, splitting, and re-sampling procedures, the models were trained on a balanced dataset of 3040 data points distributed as follows: Strong-blokers: 760, Moderate-blockers: 760, Weak-blockers: 760, and Non-blocker: 761. The reported statistics were validated on a validation set of 391 observations distributed as follows: Strong-blockers: 98, Moderate-blockers: 98, Weak-blockers: 98, Non-blocker: 97. The results on the table report the multiclass accuracy, denoted as $AC_{mul}$, followed by the binary classification metrics where the binary accuracy is denoted as $AC_{bin}$. The binary classification cut-off was based on a threshold of $PIC_{50} = 4.5$, which is equivalent to an $IC_{50} = 30\mu M$.

Table 5 illustrates that ReLu activation function provides better performing convergence compared to sigmoid in terms of accuracy. For both activation functions, we notice that the choice of batch normalization without dropout gives better performance. Concerning the best performing model, we see that models 5 and 6 have comparable binary classification (82%) and F1-score (87%); however, model 5 performs better in multiclass classification. Hence, the best MLP model of choice for our problem is model 5. This decision is further supported by the evaluation on the external test set, EV-Set-Nav, in Table 6, where model 5 scores higher in both binary and multiclass accuracies as well as in the F1-score, with 3%, 4%, and 2% improvement respectively.

**Table 5:** Performance of each model in the hyperparameter search space of the four parameters (Activation Function, Dropout Rate, Batch Normalization, and Batch Size) on the 10% validation set. The table reports the multiclass accuracies ($AC_{mul}$) and the binary metrics starting from the binary accuracy ($AC_{bin}$) onward. The binary classification is based on the threshold of $IC_{50} = 30\mu M$. Models in bold (Model 5 and 6) represent the best performing models. In 'Dropout Rate' and 'Batch Normalizations' columns, '-' means that the model was trained without the use of this hyperparameter.

| Model # | Activation Function | Dropout Rate | Batch Normalization | Batch Size | $AC_{mul}$ | $AC_{bin}$ | SN | SP | F1 | MCC | TP | FN | TN | FP |
|---|---|---|---|---|---|---|---|---|---|---|---|---|---|---|
| 1 | ReLu | 0.5 | Yes | 256 | 62.1 | 79.3 | 78.6 | 81.4 | 85.1 | 53.9 | 231 | 63 | 79 | 18 |
| 2 | ReLu | 0.5 | Yes | 512 | 61.5 | 78.3 | 78.6 | 77.3 | 84.5 | 50.5 | 231 | 63 | 75 | 22 |
| 3 | ReLu | 0.5 | - | 256 | 63.6 | 80.6 | 82.0 | 76.3 | 86.4 | 53.7 | 241 | 53 | 74 | 23 |
| 4 | ReLu | 0.5 | - | 512 | 62.0 | 74.9 | 76.9 | 69.1 | 82.2 | 41.7 | 226 | 68 | 67 | 30 |
| **5** | **ReLu** | **-** | **Yes** | **256** | **70.4** | **82.9** | **82.3** | **84.5** | **87.8** | **60.8** | **242** | **52** | **82** | **15** |
| **6** | **ReLu** | **-** | **Yes** | **512** | **69.4** | **82.4** | **82.3** | **82.5** | **87.5** | **59.2** | **242** | **52** | **80** | **17** |
| 7 | ReLu | - | - | 256 | 69.6 | 81.8 | 81.6 | 82.5 | 87.1 | 58.3 | 240 | 54 | 80 | 17 |
| 8 | ReLu | - | - | 512 | 68.5 | 76.7 | 78.2 | 72.2 | 83.5 | 45.9 | 230 | 64 | 70 | 27 |
| 9 | Sigmoid | 0.5 | Yes | 256 | 51.2 | 72.4 | 71.8 | 74.2 | 79.6 | 40.6 | 211 | 83 | 72 | 25 |
| 10 | Sigmoid | 0.5 | Yes | 512 | 50.6 | 72.1 | 71.4 | 74.2 | 79.4 | 40.3 | 210 | 84 | 72 | 25 |
| 11 | Sigmoid | 0.5 | - | 256 | 53.7 | 76.2 | 76.2 | 76.3 | 82.8 | 47.0 | 224 | 70 | 74 | 23 |
| 12 | Sigmoid | 0.5 | - | 512 | 55.1 | 75.4 | 76.2 | 73.2 | 82.4 | 44.4 | 224 | 70 | 71 | 26 |
| 13 | Sigmoid | - | Yes | 256 | 65.1 | 81.9 | 81.6 | 83.4 | 87.0 | 59.0 | 240 | 54 | 81 | 16 |
| 14 | Sigmoid | - | Yes | 512 | 64.8 | 81.3 | 80.6 | 83.5 | 86.7 | 57.9 | 237 | 57 | 81 | 16 |
| 15 | Sigmoid | - | - | 256 | 83.9 | 80.8 | 80.6 | 81.4 | 96.3 | 56.3 | 237 | 57 | 79 | 18 |
| 16 | Sigmoid | - | - | 512 | 63.5 | 80.6 | 81.0 | 79.4 | 86.2 | 55.0 | 238 | 56 | 77 | 20 |

**Table 6:** Performance of the comparable models 5 and 6 on the test set. Model 5 still proves to be the optimal

| Model # | $AC_{mul}$ | $AC_{bin}$ | SN | SP | F1 | MCC | TP | FN | TN | FP |
|---|---|---|---|---|---|---|---|---|---|---|
| **5** | **69.6** | **79.8** | **81.4** | **58.3** | **88.2** | **24.6** | **131** | **30** | **07** | **05** |
| 6 | 65.4 | 76.3 | 78.9 | 41.7 | 86.1 | 12.5 | 127 | 34 | 05 | 07 |

*4.2.2.2. RF Analysis*

This section analyses the performance of Random Forest in our development set. Here, we used RF to model a multiclass inducer. There are two hyperparameters to tune for this algorithm: the number of estimators and the maxim depth. By applying a 10-fold stratified cross validation training on a SMOTE over-sampled Dev-Set-Nav-Red, we performed a grid-based hyperparameter search to find the optimal number of decision trees of our best performing RF. In our strategy, we applied what is called by pre-pruning. The search space consisted of a set of 10 different values starting at 10 and incrementing by 10 for each next number of estimators to evaluate. The selection process was based on three metrics: the multiclass (Multiclass $AC_{cv}$) and binary (Binary $AC_{cv}$) cross-validation estimate of accuracy as well as the cross-validation estimate of F1-score ($F1_{cv}$). Those metrics are derived by computing the mean value of the accuracies achieved at each validation fold during one complete CV run and using the same number of estimators. The results are displayed in Table S1 under the Supplementary Information section.

We noticed that when the number of estimators is above 30, we achieve almost the same Binary $AC_{cv}$ with similar $F1_{cv}$. Concerning the Multiclass $AC_{cv}$, we noticed that it keeps increasing slowly while increasing the number of estimators but still stays within a comparable range. Therefore, we chose the best performing Random Forest with the least number of estimators, i.e. 30 number of estimators, as a form of regularization. The next step is to find the best maximum depth of the trees using an ensemble of 30 estimators. The hyperparameter search is performed again via a 10-fold stratified CV. Table S2, in the Supplementary Information section, shows the results of the search.

The statistics are arbitrary close to each other, in Table S2. However, a max depth of 24 shows the highest performance (in terms of fold multiclass-accuracy, binary-accuracy, and F1-score). At the same time, it is the most commonly found best max depth in the 10-fold stratified CV. Therefore, a bootstrapping-based RF with 30 estimators and a max depth of 24 will be used to build our optimal RF multiclass model on the entire re-sampled Dev-Set-Nav-Red.

*4.2.2.3. Kernelized SVM Analysis*

To analyze the performance of kernelized SVMs in this problem, we followed a one-vs-rest approach to classify molecular compounds at all of the three potency levels. We applied grid search on the constrained convex objective function defined in eq. 5 to identify the best parameters of the optimal kernelized SVMs. The grid-based search, which was conducted following a 10-fold stratified cross validation, counted 3 hyperparameters to examine:

- *Kernel function:* 4 kernels including the linear, polynomial(poly), sigmoid, and radial basis function(rbf) kernel
- *Polynomial degree:* In case of polynomial kernel, 9 degrees were examined (from 2 to 10)
- *Penalty constant C:* For constraints relaxation/penalization, 10 values were evaluated of the penalty constant C, with 5 values less or equal than 1 (0.1,0.2,0.5,0.8, 1) denoting a soft margin SVM and 5 greater than 1 (3, 5, 10, 50, 100) denoting hard margin SVM

The search space resulted in 120 different models for each of the original and re-sampling methods, and at each of the three potency thresholds. In total, to build our final classifier, we had to evaluate 3x3x120 = 1080 kernelized SVM models. Each row in Table 7 corresponds to the best performing model among the 120 models evaluated at a given threshold and using one of the sampling techniques. Each row shows the best model's $Ac_{cv}$ and the $F1_{cv}$ along with the model configuration parameters.

In Table 7, we see that the radial basis function(rbf) represents the best performing kernel for our given problem. For both thresholds of $PIC_{50} = 6$ and $PIC_{50} = 4.5$, performance results between training on the imbalanced/original datasets and balanced ones does not show that we are losing much by not adding synthetic samples. We are already performing at a level above 85% for accuracy and 75% for F1-score. The radial basis function seems finding a better generalizing decision boundary, with a quite hard margin, on the original sets. Hence, we are opting for 6svm and 4o5svm as the optimal models for our model building at this potency levels. Concerning the threshold of $PIC_{50} = 5$, training on the original dataset does not seem to generalize well enough compared to a SMOTE over-sampled training set. We see an increase by ~6% when adding synthetic samples to the training set.

**Table 7:** The best performing model, among the 120 models evaluated, at each given threshold and using one of the sampling techniques. In '*Degree*' column, '-' represents the model does not take a degree parameter

| $PIC_{50}$ Threshold | Model short name | Sampling strategy | Development class distribution | | $Ac_{cv}$ | $F1_{cv}$ | Kernel | Penalty Constant | Degree |
|---|---|---|---|---|---|---|---|---|---|
| | | | blk | nblk | | | | | |
| 4.5 | **4o5svm** | **Original** | **1442** | **108** | **87.8** | **88.6** | **rbf** | **5** | **-** |
| | 4o5svm-ovrs | Over-sampling | 1442 | 1442 | 88.2 | 88.1 | rbf | 3 | - |
| | 4o5svm-unds | Under-sampling | 108 | 108 | 80.3 | 78.8 | rbf | 10 | - |
| 5 | 5svm | Original | 1018 | 532 | 76.7 | 80 | poly | 0.2 | 3 |
| | **5svm-ovrs** | **Over-sampling** | **1018** | **1018** | **82.5** | **83.1** | **rbf** | **10** | **-** |
| | 5svm-unds | Under-sampling | 532 | 532 | 77.8 | 77.3 | rbf | 100 | - |
| 6 | **6svm** | **Original** | **160** | **1390** | **85.6** | **75.3** | **rbf** | **5** | **-** |
| | 6svm-ovrs | Over-sampling | 1390 | 1390 | 88.5 | 88.5 | rbf | 3 | - |
| | 6svm-unds | Under-sampling | 532 | 532 | 79.3 | 79.7 | rbf | 5 | - |

Also, the class distribution is not skewed, where blockers represent double the molecular compounds of non-blockers with a fair number of total observations. Therefore, we opted for SMOTE re-sampling to bias the generalization process of moderate blockers predictor. We chose 5svm-ovrs as the best optimal model at this level. The selected models were also evaluated individually on EV-SET-NAV and proven to be highly performant on unseen data, as shown in Table 8.

**Table 8:** Prediction performance of best individual kernelized SVM models on the evaluation set

| Model | AC | SN | SP | F1 | TP | FN | TN | FP |
|---|---|---|---|---|---|---|---|---|
| 4o5svm | 85 | 85 | 83 | 91 | 137 | 24 | 10 | 02 |
| 5svm-ovrs | 83 | 86 | 80 | 87 | 98 | 16 | 47 | 12 |
| 6svm | 83 | 72 | 84 | 47 | 13 | 05 | 131 | 24 |

For our final multiclass classifier, we organized the one-vs-rest Nav1.5 predictive sub-models in a form of a binary tree, where inducers are applied sequentially, to filter Nav1.5 blockers from non-blockers at the three potency thresholds. The final model (visualized in Figure 10), is denoted as ToxTree-Nav1.5.

A new test molecule with unknown Nav1.5 potency will be evaluated by ToxTree-Nav1.5 model as follows: Given the extracted 2D descriptors using PaDEL-descriptors, we first, extract the relevant 2D descriptors, normalize those features, and then apply a linear transformation to perform dimensionality reduction. Once, we have the final 22-dimensional vector embedding of the compound, the three selected best performing SVM models (6svm, 5svm-ovrs, and 4o5svm) are applied sequentially in the given order. If the molecular compound is filtered out by one model as a non-blocker, it will be passed on to the following model, for further evaluation, in the pipeline. However, if the compound is predicted as a Nav1.5 blocker, it will be described as either 'Strong blocker', 'Moderate Blocker', or 'Weak blocker' according to the model that took the decision in the order of the pipeline respectively.

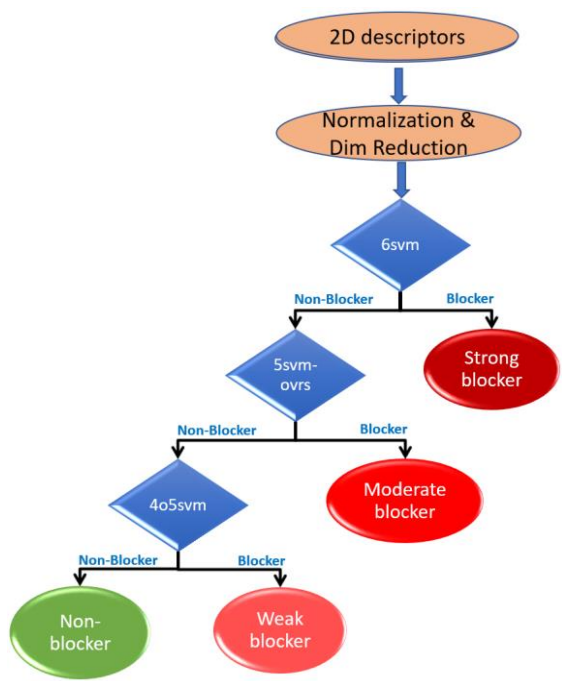

**Figure 10:** ToxTree-Nav1.5 Classifier conceptual visualization inference pathway

### 4.2.3. ToxTree-Nav1.5 Regressor

Similarly to hERG advanced data analytics, we have also built a regression model, on the 1550 unique set of molecular compounds, predicting their $PIC_{50}$ potency value. The classification performance of the regression model, ToxTree-Nav1.5 Regressor, was then compared to the one of multiclass inducer, ToxTree-Nav1.5 Classifier, on the same test set (section 4.2.4).

A ridge regression model was compared to a random forest regressor. The random forest regressor achieved higher performance on the cross-validation estimate of MSE. The best model achieved an MSE = 0.26 on the evaluation set of N=173.

Figure S7 in the supplementary information represents the performance of our regressor in predicting the $PIC_{50}$ potency values of molecular compounds with respect to the ground truth values. The model is denoted as ToxTree-Nav1.5 Regressor in the rest of the paper and achieves a coefficient of determination $r^2 = 0.71$ on the test set.

*4.2.4. Comparative Prediction Performance of the Four Nav1.5 Models*

While comparing performances of the four Nav1.5 models (MLP, RF, ToxTree-Nav1.5 Regressor, and ToxTree-Nav1.5 Classifier), we notice that the one-vs-rest approach is outperforming the other three in the binary classification task. Concerning ToxTree-Nav1.5 Regressor, the performance results were better than expected. The composite descriptors created by the linear transformation step captured well the embedded information of our dataset. All of the four models were compared on the same test set of N=173 molecular compounds labeled at the threshold $PIC_{50} = 5$.

The observation on the one-vs-rest approach can also be derived from the performances reported in Table 7, where we see the binary cross validation estimate of accuracy of the three selected models scoring as high as 85% with higher F1-scores. Furthermore, the sub-models selected to build the final inducer were trained mainly on original labeled datasets compared to the RF and MPL that were trained on SMOTE re-sampled datasets to achieve their best performances. Therefore, our best final deployed model will be ToxTree-Nav1.5 Classifier. This decision is further supported by the evaluation of the three models on the unseen test set EV-Set-Nav (N=173). Table 9 confirms the final decision of our best model. The performance of ToxTree-Nav1.5 Classifier shows an improvement of 4% to 6% for binary classification and higher performance in the rest of the binary metrics. Concerning multiclass classification, our selected final model demonstrates an out-performance to both the Random Forest and the deep learning model with an improvement of more than 4%.

**Table 9:** Comparing prediction performance of ToxTree-Nav1.5 with MLP and RF best models on the unseen test set EV-Set-Nav

| Inducer | *Multi*AC | AC | CCR | MCC | SN | SP | F1 | TP, FN, TN, FP |
|---|---|---|---|---|---|---|---|---|
| MLP | 69.6 | 80.9 | 76.5 | 56.2 | **90.4** | 62.7 | 86.2 | 103, 11, 37, 22 |
| RF | 70.3 | 82.1 | 79.0 | 59.4 | 88.6 | 69.5 | 86.7 | 101, 13, 41, 18 |
| ToxTree-Nav1.5 Regressor | - | 75.7 | 73.8 | 46.9 | 79.8 | 67.8 | 81.2 | 91, 23, 40, 19 |
| ToxTree-Nav1.5 Classifier | **74.9** | **86.7** | **86.2** | **71.2** | 87.7 | **84.7** | **89.7** | 100, 14, 50, 09 |

## 5. Conclusion

Cardiac ion channels are a group of voltage-gated channels that function collectively and in full harmony to generate the action potential, which is needed for cardiac cells' contraction. The ionic currents produced by the hERG and Nav1.5 ion channels form a major component of the cardiac action potential and blocking these channels by small molecule drugs can lead to severe cardiovascular complications. This rising concern has pushed scientists and researchers to look for new methods to identify hERG and Nav1.5 blockers. The recent exponential increase of bioactivity data on these channels makes them well-suited to build robust machine learning models to predict their liability. Most research work in the field has focused only on one cardiac ion channel, namely hERG. Here, we describe two models to predict both hERG and Nav1.5 drug-mediated liability. Our ML models used two large manually curated datasets of size 8380 and 1550 for hERG and Nav1.5, respectively. The two sets hold potency information of the two targets and were used to build 2D descriptor-based multiclass classification inducers at three different potency cut-offs (*i.e.* 1μM, 10μM and 30μM). The first model, named ToxTree-hERG Classifier, represents an ensemble of Random Forest models applied sequentially to predict the inhibition of a new compound with unknown potency. The second model, named ToxTree-Nav1.5 Classifier, consists of a set of one-vs-rest carefully trained kernelized SVM sub-models run sequentially to filter out the potency level of a compound. The hERG model was evaluated on an external test set of N=499 compounds achieving a binary performance of Q2 = 93.2% and a multiclass accuracy of Q4=74.5%. The model was also benchmarked to existing tools and inducers showing an improvement of 1.2% to 53.2% in classifying blockers from non-blockers, beating the state-of-the-art consensus model. Whereas, the first Nav1.5 liability predictive model was evaluated on an external balanced test set of 173 unique compounds extracted from PubChem. Three different robust models were built and analysed for best final model selection, namely a MLP, a multiclass RF, and a pipeline of kernelized SVMs(ToxTree-Nav1.5 Classifier). ToxTree-Nav1.5 AI was selected as the best classifier model as most of the sub-models were trained on original datasets, without any re-sampling techniques needed. Also, the selected model beats the other two classification models in binary classification and achieves a comparable Q4 performance with RF while outperforming the MLP. The final model yielded a binary classification performance of Q2 = 86.7% and a multiclass accuracy of Q4=74.9%.

## 6. Acknowledgment

The authors would like to acknowledge funding from the Natural Sciences and Engineering Research Council of Canada (NSERC) Discovery grant.

# Supplementary Information

1. *hERG Data Collection and Curation*

As a first step to further clean the data, an exact duplicate analysis of the SMILE strings showed that a subset of 23 unique molecular compounds had exact duplicate matches. All of these duplicates were in the training set as mentioned by Kumar et al. [6], used to train their published model. Most of the time, those duplicates had different potencies leading to the assignment of the same compound to different classes, which introduced noise to the data. For duplicates falling in the same class, we kept a unique SMILE string and reassigned the potency to the average among its duplicates. In case the potencies were too far, i.e. they map to different classes, we discarded the compound. This led to a reduced set of 9178 instances. Furthermore, and using the PubChem's ID exchange service[1], we fed the entire cleaned set of SMILE strings to the server. The output results showed that a subset of 596 compounds contained duplicate entries mapping to the same CID. Out of the 277 compounds retrieved from PubChem and used for evaluation purposes by Kumar and his colleagues, 226 compounds were identified having duplicates in the training set. Second, out of the 222 compounds fetched from the literature and used as well in the test set, 7 were identified having duplicates in the training set. Third, 67 compounds in the training set had duplicate entries in the same set. In our second curation process, we removed all the duplicates from the training set (i.e. compounds retrieved from ChEMBL) and kept their corresponding unique ones in the test set. Concerning the duplicates within the training set, we followed the same manual curation procedure done in the first investigation step. The final dataset includes 8380 instances for development ($PIC_{50}$ distribution is visualized Fig, S1.a), and 499 instances used for testing ($PIC_{50}$ distribution is visualized Fig, S1.b).

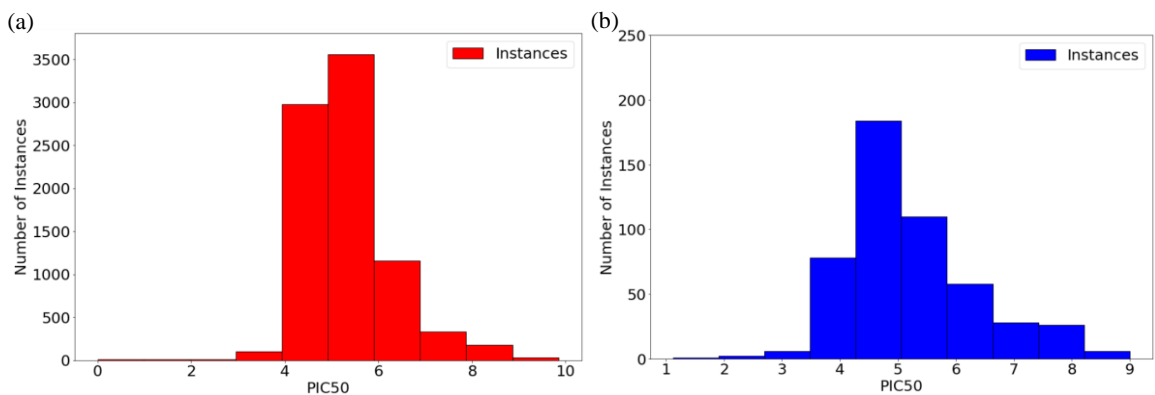

**Figure S1:** $PIC_{50}$ distribution of the (a) training and (b) test hERG liability dataset

2. *Nav1.5 Data Collection and Curation*

For redundancy reduction, we noticed that many redundant molecular compounds were reporting assays conducted on 2 cell types: Chinese Hamster Ovary (CHO) K1 and HEK293 cells. Since potency values on the two cells were too different from each other, we performed a count analysis of the most dominant cells used in the Nav1.5 inhibition assays to choose which one to consider. On the original retrieved dataset, the count represented 1078 for HEK293 and 181 for CHO. Hence, only assays on HEK293 cells were considered in case of redundancy. If we still have more than one potency value reported on the same compound, we discarded the compound from any further analysis if the values difference is greater by more than one order of magnitude. Otherwise, the mean was computed and assigned. At the end of the curation process, we ended up with 1463 unique records with reported $IC_{50}$ value through assays. Furthermore, we compared the final set we obtained with the curated ChEMBL v25 set based on 100% similarity of the compounds' SMILE strings. We compiled 18 exact similar compounds in both sets. Since we believe that ChEMBL database is better maintained, we wanted to keep entries retrieved from this database and discard the others. Therefore, we removed those 18 duplicates from PubChem curated set and kept them in ChEMBL v25 one. The final sets represent 1646 (Fig. S2) and 1437 (Fig. S3) unique compounds for ChEMBL v25 and PubChem respectively. All $IC_{50}$ values were normalized to $PIC_{50}$ values.

---

1 https://pubchem.ncbi.nlm.nih.gov/idexchange/idexchange.cgi

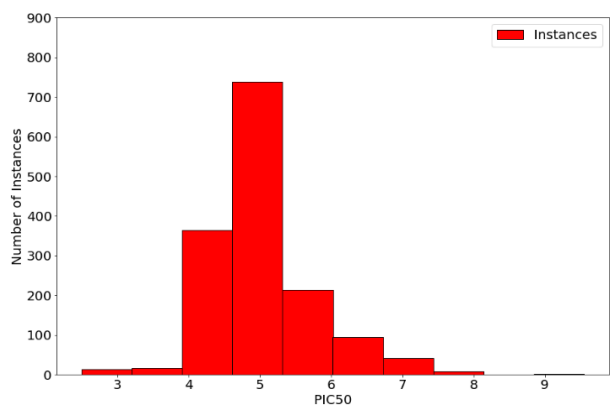
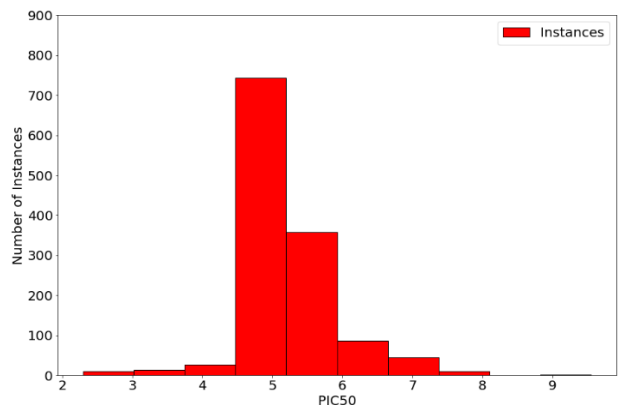

**Figure S2:** $PIC_{50}$ distribution of ChEMBL v25 curated Nav1.5 liability dataset

**Figure S3:** $PIC_{50}$ distribution in PubChem curated Nav1.5 liability dataset

The final goal of this work was to be able to classify Nav1.5 blockers and non-blockers. For the blocking molecular compounds, we wanted to achieve a multiclass classification at three intensity levels: strong, moderate, and weak blockers at three potency thresholds, namely at a $PIC_{50}$ value of 6, 5, and 4.5 respectively. An analysis of the ChEMBL retrieved set conveyed that out of 1646 instances, only 86 compounds had $PIC_{50} < 4.5$ (for non-blockers) and 170 compounds with $PIC_{50} \geq 6$ (for strong blockers). Whereas, the 1437 PubChem curated set comprised of 88 compounds with $PIC_{50} < 4.5$ (for non-blockers) and 145 compounds with $PIC_{50} \geq 6$ (for strong blockers).

To augment our dataset, especially for the minority classes, we appended all the 88 non-blockers and the 145 strong blockers from PubChem to the cleaned ChEMBL set. Before moving to the model development phase, we fed the gathered set of SMILE strings to the PubChem's ID exchange service, in order to check for potential duplicate compounds that have different SMILE representations, yet similar in structure. The redundancy reduction procedure reduced the gathered data by 156 instances, to leave us with a final high-quality set of 1723 compounds. We then applied a stratified split based on the 4 classes previously explained to select 10% (173 molecular compounds) for evaluation and 90% (1550 molecular compounds) for development. The final dataset includes 1550 instances for development ($PIC_{50}$ distribution is visualized Fig, S4.a), and 173 instances used for testing ($PIC_{50}$ distribution is visualized Fig, S4.b).

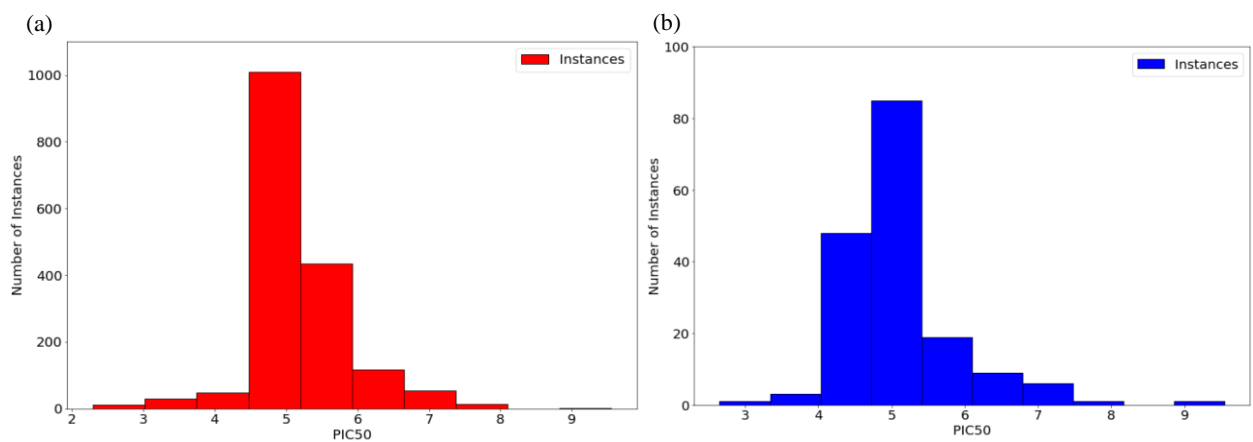

**Figure S4:** $PIC_{50}$ distribution of the (a) training and (b) test Nav1.5 liability dataset

3. *Nav1.5 feature selection – percent missing value and information analysis*

Through a single feature analysis of the data, we noticed that many columns had most (~71%) of their values as the same constant, namely a value of zero. Such features contain no information and will just add noise to our model. Therefore, features with a threshold of more than 1300 zero values were discarded from any further analysis, resulting in a reduction of 386 features to be removed. Additionally, we noticed that many features had *Nan* values, some features with more than 600 *Nan* values and many with more than 210 *Nans*. Assigning a value of zero or a mean value to those statistics will lead to biased models. A threshold of more than 200 *Nan* values per column was set to discard those features. This feature reduction procedure resulted in 551 features to be considered for model building.

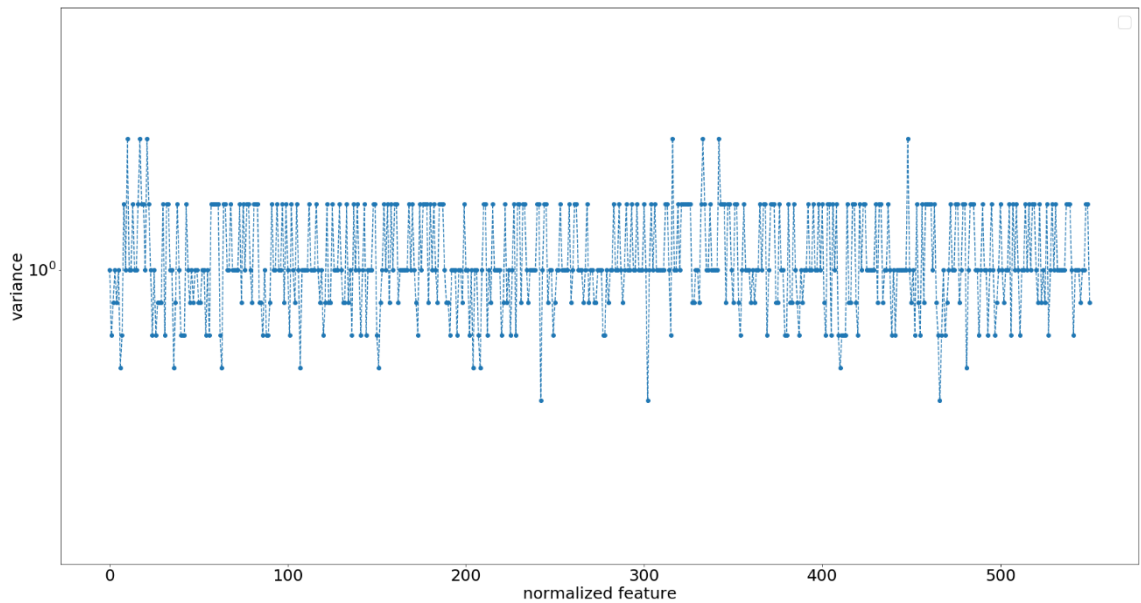

**Figure S5:** Dev-Set-Nav variance-based analysis of the normalized 551 features

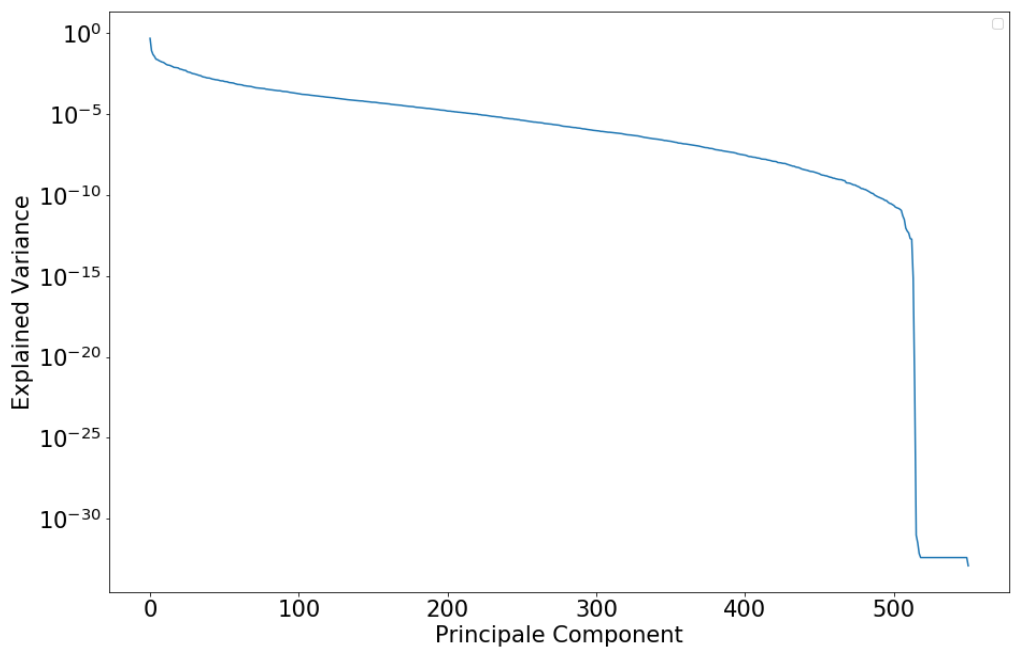

**Figure S6:** Principal component analysis of the 551 features of the entire normalized Dev-Set-Nav where x-axis represents the principal components and the y-axis their corresponding variance

**Table S1:** Number of estimators' selection based on the performance of each random forest with an ensemble of decision trees using 10-fold stratified cross-validation. The optimal number of estimators is highlighted in bold.

| Number of estimators | Multiclass $AC_{cv}$ | Binary $AC_{cv}$ | $F1_{cv}$ |
|---|---|---|---|
| 10 | 69.7 | 81.8 | 81.5 |
| 20 | 71.2 | 82.3 | 80.9 |
| **30** | **71.5** | **82.4** | **81** |
| 40 | 71.8 | 82.6 | 81 |
| 50 | 71.8 | 82.4 | 80.9 |
| 60 | 71.9 | 82.5 | 81 |
| 70 | 71.9 | 82.5 | 81 |
| 80 | 72.1 | 82.5 | 81 |
| 90 | 72.1 | 82.5 | 81 |
| 100 | 72.1 | 82.4 | 81 |

**Table S2:** Max depth selection based on the performance of each random forest with the same best ensemble of decision trees using 10-fold stratified cross-validation. The optimal max depth is highlighted in bold.

| Model number | Number of estimators | Max depth | Fold multiclass-accuracy | Fold binary-accuracy | Fold F1-score |
|---|---|---|---|---|---|
| RF_1 | 30 | 23 | 70.4 | 82.7 | 81.1 |
| RF_2 | 30 | 23 | 72.3 | 82.9 | 81.3 |
| **RF_3** | **30** | **24** | **70** | **82.7** | **81** |
| RF_4 | 30 | 25 | 70.9 | 82.9 | 81.3 |
| RF_5 | 30 | 26 | 71.9 | 82.4 | 80.9 |
| RF_6 | 30 | 27 | 71.1 | 82.7 | 81.1 |
| RF_7 | 30 | 22 | 71.9 | 82.1 | 80.7 |
| RF_8 | 30 | 25 | 70.9 | 81.6 | 80.3 |
| **RF_9** | **30** | **24** | **70.6** | **81.3** | **80.2** |
| **RF_10** | **30** | **24** | **75.2** | **83.2** | **81.5** |

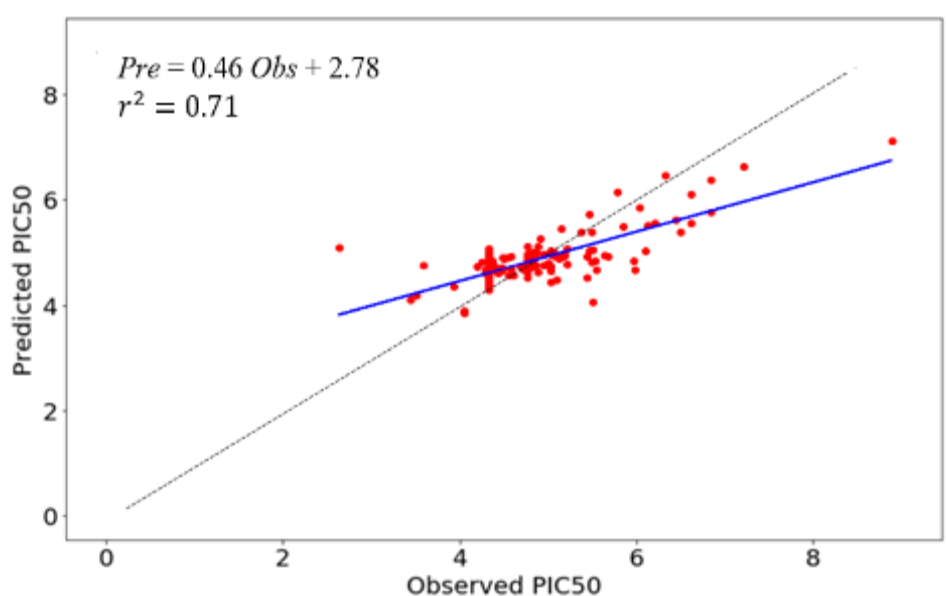

**Figure S7:** Regression scatter plot of the Predicted vs. Observed $PIC_{50}$ of our ToxTree-Nav1.5 Regressor model on the evaluation dataset (EV-Set-Nav) of N=173 unique molecular compound